\documentclass[runningheads,a4paper]{llncs}

\usepackage{amssymb}
\usepackage{amsmath}
\usepackage{graphicx}
\usepackage{subfig}
\usepackage{epsfig}
\usepackage{url}
\usepackage{booktabs}
\usepackage{tabularx}
\usepackage{xcolor}
\usepackage{listings}
\usepackage[colorlinks,bookmarks=false]{hyperref}
\expandafter\def\expandafter\UrlBreaks\expandafter{\UrlBreaks\do\_\do\-}

\AtBeginDocument{}

\begin{document}

\mainmatter  % start of an individual contribution

\title{Semantic Context Forests for Learning-Based Knee Cartilage Segmentation in 3D MR Images}

\titlerunning{Learning-Based Knee Cartilage Segmentation}

\author{Quan Wang$^{\star,\dagger}$ \and
Dijia Wu$^\star$ \and Le Lu$^\star$ \and Meizhu Liu$^\star$ \and \\ Kim L. Boyer$^\dagger$ \and Shaohua Kevin Zhou$^\star$}
%\author{******************************************************************\\
%******************************************************************}
%
\authorrunning{Q. Wang \and D. Wu \and L. Lu \and M. Liu \and K. L. Boyer \and S. K. Zhou}
%\authorrunning{*********************************}

\institute{$^\star$Siemens Corporate Research, Princeton, NJ 08540, USA\\
$^\dagger$Rensselaer Polytechnic Institute, Troy, NY 12180, USA}
%\institute{******************************************************************\\
%******************************************************************}


\toctitle{Learning-Based Knee Cartilage Segmentation}
\tocauthor{Quan Wang, \textit{et al.}}
\maketitle
\vspace*{-5.5mm}
\begin{abstract}
The automatic segmentation of human knee cartilage from 3D MR images is a useful yet challenging task due to the thin sheet structure of the cartilage with diffuse boundaries and inhomogeneous intensities. In this paper, we present an iterative multi-class learning method to segment the femoral, tibial and patellar cartilage simultaneously, which effectively exploits the spatial contextual constraints between bone and cartilage, and also between different cartilages. First, based on the fact that the cartilage grows in only certain area of the corresponding bone surface, we extract the distance features of not only to the surface of the bone, but more informatively, to the densely registered anatomical landmarks on the bone surface. Second, we introduce a set of iterative discriminative classifiers that at each iteration, probability comparison features are constructed from the class confidence maps derived by previously learned classifiers. These features automatically embed the semantic context information between different cartilages of interest.
Validated on a total of 176 volumes from the Osteoarthritis Initiative (OAI) dataset, the proposed approach demonstrates high robustness and accuracy of segmentation in comparison with existing state-of-the-art MR cartilage segmentation methods.

\textbf{Code: } {\url{https://github.com/wq2012/kneeseg}}

\textbf{Pretrained Models: } {\url{https://huggingface.co/collections/wq2012/knee-mri-segmentation}}

%\keywords{Knee cartilage segmentation, spatial contextual constraints, iterative discriminative classifiers, OAI dataset}

\end{abstract}

\vspace{-10mm}
\section{Introduction}
\vspace{-2mm}
The quantitative analysis of knee cartilage is advantageous for the study of cartilage morphology and physiology. In particular, it is an important prerequisite for the clinical assessment and surgical planning of the cartilage diseases, such as knee osteoarthritis which is characterized as the cartilage deterioration and a prevalent cause of disability among elderly population. As the leading imaging modality used for articular cartilage quantification \cite{QAOC-2004}, magnetic resonance (MR) imaging provides direct and noninvasive visualization of the whole knee joint including the soft cartilage tissues (Fig. \ref{fig:anatomy2}). However, automatic segmentation of the cartilage tissues from MR images, which is required for accurate and reproducible quantitative cartilage measures, still remains an open problem because of the inhomogeneity, small size, low tissue contrast, and shape irregularity of the cartilage.

An earlier endeavor on this problem is Folkesson \textit{et al.}'s voxel classification approach \cite{voxel_classification}, which runs an approximate $k$NN classifier on voxel intensity and absolute position based features. However, due to the overlap of intensity distribution between cartilage and other tissues such as menisci and muscles, as well as the variability of the cartilage locations from scan to scan, the performance of this method is limited. More recently, Vincent \textit{et al.} have developed a knee joint segmentation approach based on active appearance model (AAM), which captures the statistics of both object shape and image cues. Though promising results are reported in \cite{FASO-2010}, the search for the initial model pose parameter can be very time consuming even if a coarse to fine searching strategy is used.

Given the strong spatial relation between the cartilages and bones in the knee joint, most proposed cartilage segmentation methods are based on a framework that each bone is segmented first in the knee joint \cite{fripp,logismos,opt_prob}, which is usually easier than direct cartilage segmentation because the bones are much larger in size with more regular shapes. Fripp \textit{et al.} segment the bones based on 3D active shape model (ASM) incorporating the cartilage thickness statistics, and the outer cartilage boundary is then determined by examining the intensity profile along the normal to the bone surface, while being constrained by the cartilage thickness model \cite{fripp}. In Yin's work \cite{logismos}, the volume of interest containing the bones and cartilages is first detected using a learning-based approach, then the bones and cartilages are jointly segmented by solving an optimal multi-surface detection problem via multi-column graph cuts \cite{multicolumn_graph_cut}. Lee \textit{et al.} employ a constrained branch-and-mincut method with shape priors to obtain the bone surface, and then segment the cartilage with MRF optimization based on local shape and appearance information \cite{opt_prob}. In spite of the differences, these approaches all require classification of bone surface voxels into bone cartilage interface (BCI) and non-BCI, which is an important intermediate step to determine the search space or impose prior constraint for cartilage segmentation. Therefore, any classification error of BCI will probably propagate to the final cartilage segmentation result.

In this paper, we present a fully automatic learning-based voxel classification method for cartilage segmentation. It also requires pre-segmentation of corresponding bones in the knee joint. However, the new approach does not rely on explicit classification of BCI. Instead, we construct distance features from each voxel to a large number of anatomical landmarks on the surface of the bones to capture the spatial relation between the cartilages and bones. By removing the intermediate step of BCI extraction, the whole framework is simplified and classification error propagation can be avoided.

Besides the connection between the cartilages and bones, strong spatial relation also exists among different cartilages which is more often overlooked in earlier approaches. For example, the femoral cartilage is always above the tibial cartilage and two cartilages touch each other in the region where two bones slide over each other during joint movements. To utilize this constraint, we introduce the iterative discriminative classification that at each iteration, the multi-class probability maps obtained by previous classifiers are used to extract semantic context features. In particular, we compare the probabilities at positions with random shift and compute the difference. These features, which we name as the random shift probability difference (RSPD) features, are more computationally efficient and more flexible for different range of context compared to the calculation of probability statistics at fixed relative positions \cite{auto-context,ipmi2011}.

\vspace{-2mm}
\section{Review of Bone Segmentation}
\vspace{-2mm}
\label{sec:2}
In this work, we employ a learning-based bone segmentation approach which has shown the efficiency and effectiveness in different medical image segmentation problems \cite{HLBA-2008,FCHM-2008}.
We represent the shape of a bone by a closed triangle mesh $\mathcal{M}$. Given a number of training volumes with manual bone annotations, we use the coherent point drift algorithm (CPD) \cite{CPD} to find anatomical correspondences of the mesh points and thereof construct the statistical shape models with mean shape $\overline{\mathcal{M}}$ \cite{ASM}.
As shown in Fig. \ref{fig:bone_seg}, the whole bone segmentation framework comprises three steps.

\begin{figure}
  \centering
  \includegraphics[width=0.95\textwidth]{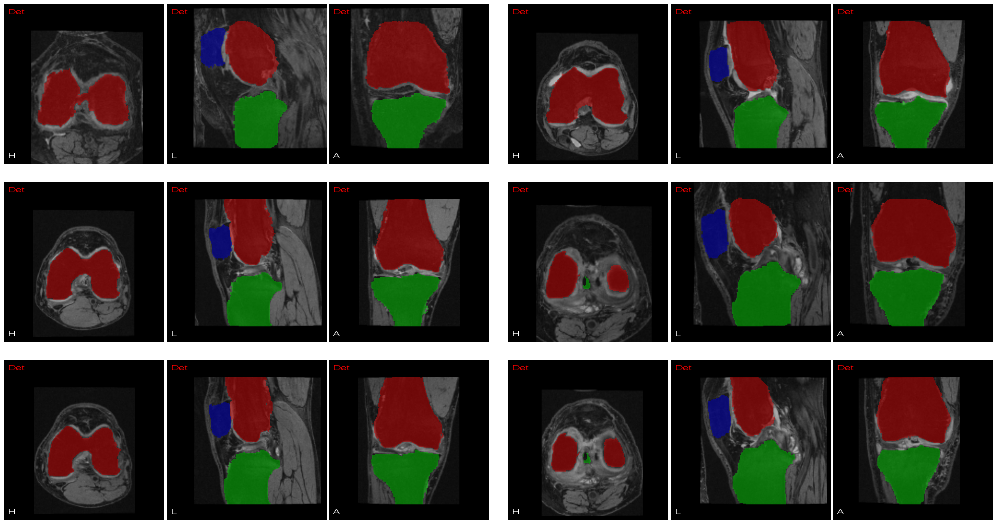}
  \caption{
  Example bone segmentations. Each case has three views, from left to right: transversal plane, sagittal plane, coronal plane. Red: femur; green: tibia; blue: patella. }
  \label{fig:bone_result}
  \vspace{-3mm}
  \end{figure}

\begin{enumerate}
\item{Pose Estimation: For a volume $\mathcal{V}$, the bone is first localized by searching for the (sub-)optimal pose parameters $(\hat{t},\hat{r},\hat{s})$, \textit{i.e.}, the translation, rotation and anisotropic scaling, using  the marginal space learning (MSL) \cite{FCHM-2008}:
\vspace{-2mm}
\begin{equation}
(\hat{t},\hat{r},\hat{s}) \approx  (\arg \max_{t}P(t|\mathcal{V}), \arg \max_{r}P(r|\mathcal{V},\hat{t}), \arg \max_{s}P(s|\mathcal{V},\hat{t},\hat{r})) ,
\vspace{-3mm}
\end{equation}
and the shape is initialized by linearly transforming the mean shape $\overline{\mathcal{M}}$. }
\item{Model Deformation: At this stage, the shape is repeatedly deformed to fit the boundary and projected to the variation subspace until convergence.}
\item{Boundary Refinement: To further improve the segmentation accuracy, we use the random walks algorithm \cite{random_walk} to refine the bone boundary (see Table \ref{tab:dsc_bone} and Fig. \ref{fig:bone_result} for results)
and employ the CPD algorithm to obtain anatomically equivalent landmarks on the refined bone surface. }
\end{enumerate}

\vspace{-2mm}
\section{Cartilage Classification}
\vspace{-2mm}
\label{sec:3}
Given all three knee bones being segmented, we first extract a band of interest within a maximum distance threshold from each of the bone surface, and only classify voxels in the band of interest to simplify the training and testing by removing irrelevant negative voxels.

\subsection{Feature Extraction}

For each voxel with spatial coordinate $\mathbf{x}$, we construct a number of base features which can be categorized into three subsets.

\textbf{Intensity Features} include the voxel intensity and its gradient magnitude, respectively:
$f_1(\mathbf{x})=I(\mathbf{x})$, $f_2(\mathbf{x})=||\nabla I(\mathbf{x})||$.

\begin{figure}[h!]
  \centering
  \subfloat[]
  {   \label{fig:d2lm}
      \includegraphics[height=0.32\textwidth]{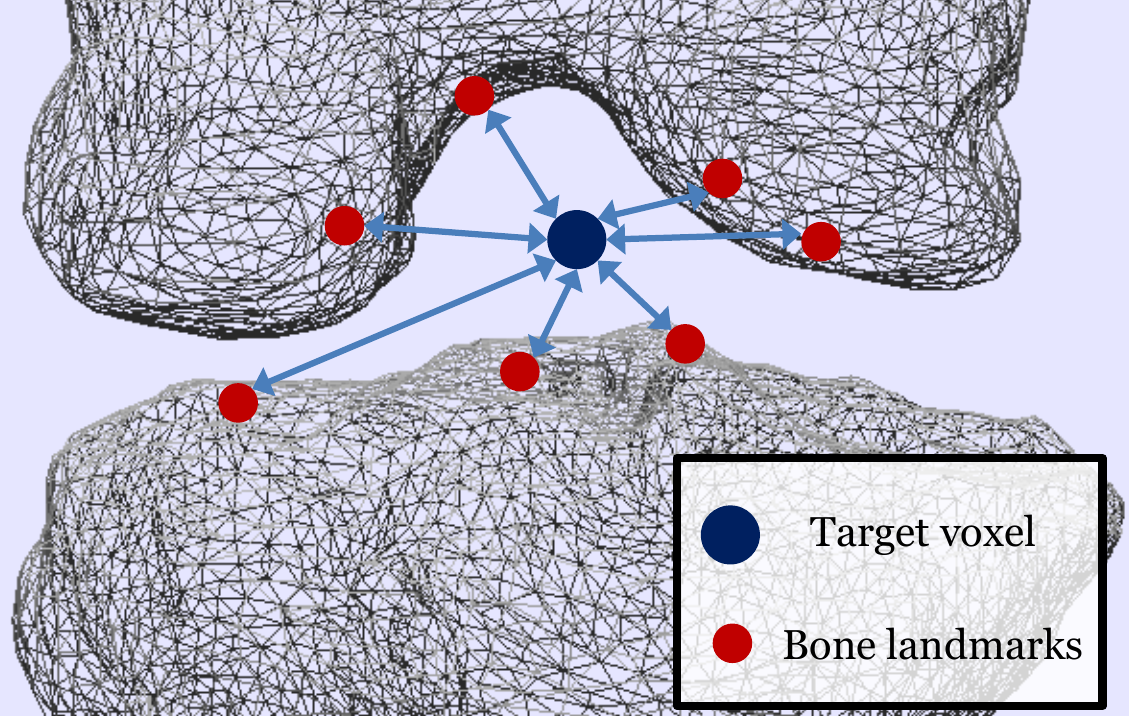}
  }\;
  \subfloat[]
  {   \label{fig:feature_scatter}
      \includegraphics[height=0.32\textwidth]{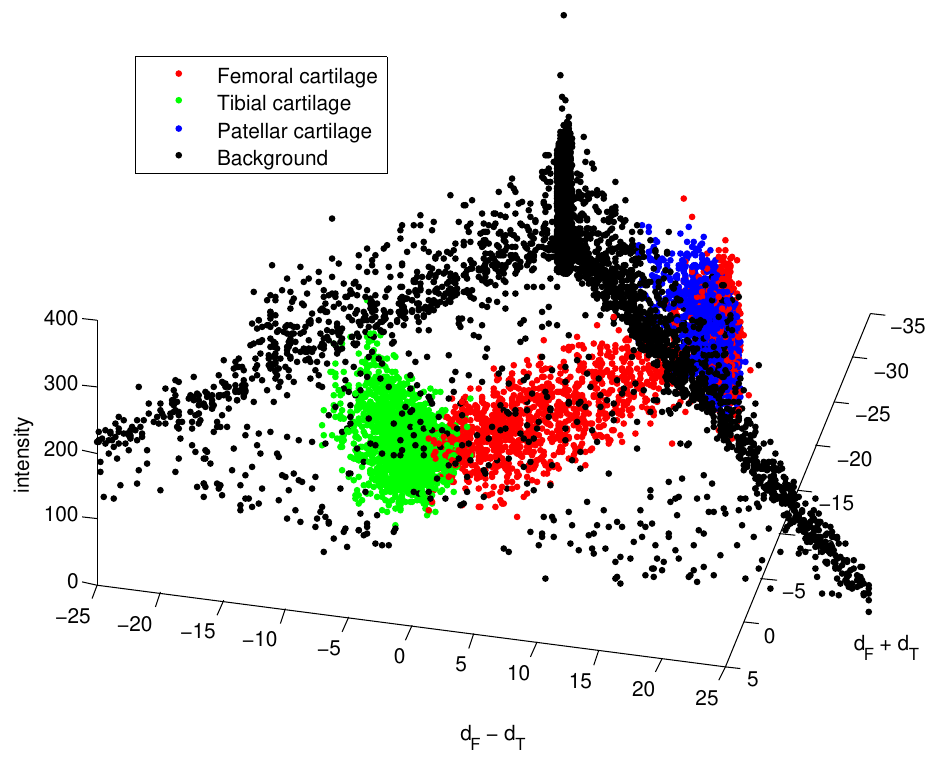}
  }\\
  \subfloat[]
  {
      \label{fig:feature_frequency}
      \includegraphics[height=0.28\textwidth]{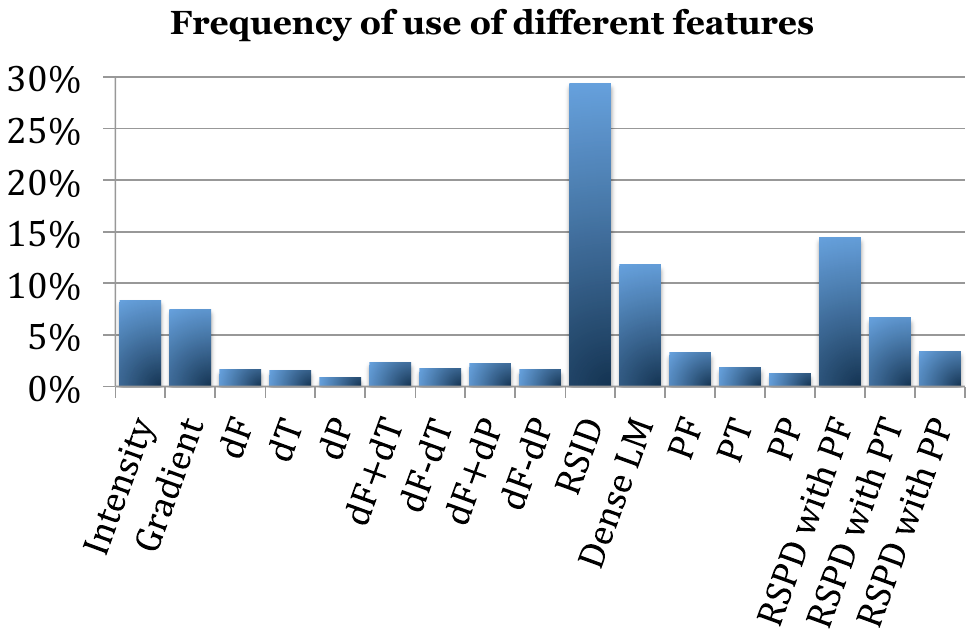}
  }\;
  \subfloat[]
  {
      \label{fig:bar}
	  \includegraphics[height=0.28\textwidth]{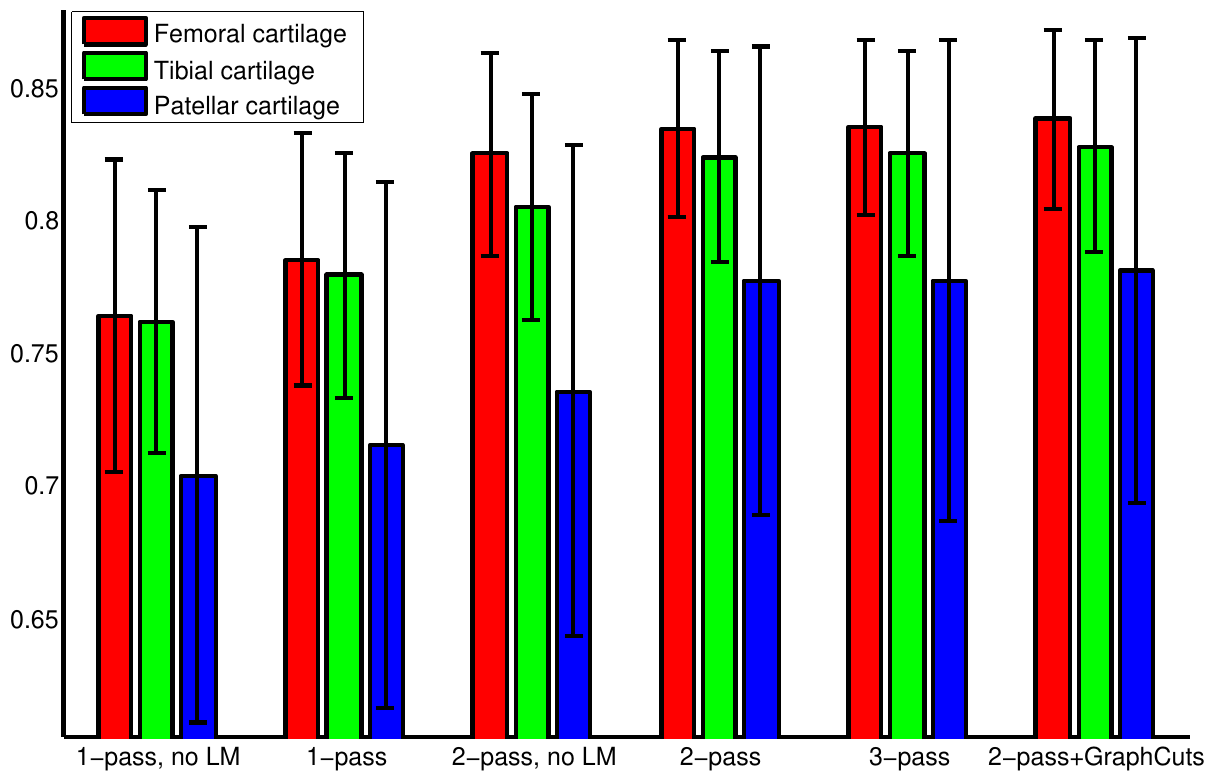}
  }
  \caption{
  (a) Distances to densely registered bone landmarks encode anatomical position of a voxel.
  (b) Feature scatter plot: intensity and distance features separate tibial cartilage from femoral and patellar cartilages.
  (c) Frequency of each feature selected by the classifier in the 2nd pass.
  (d) A comparison of segmentation performance (DSC): 1-pass/2-pass forests without using distance to landmark (LM) features; 1-pass/2-pass/3-pass forests using distance to landmark features; 2-pass forests with graph cuts optimization (3-fold cross validation).  }
\end{figure}

\textbf{Distance Features} measure the signed Euclidean distances from each voxel to different knee bone boundaries:
$f_3(\mathbf{x})=d_F(\mathbf{x})$, $f_4(\mathbf{x})=d_T(\mathbf{x})$,
$f_5(\mathbf{x})=d_P(\mathbf{x})$,
where $d_F$ is the signed distance to the femur, $d_T$ to tibia, and $d_P$ to patella.
Then we have their linear combinations:
\vspace{-2mm}
\begin{equation}
f_{6/7}(\mathbf{x})=d_F(\mathbf{x}) \pm d_T(\mathbf{x}),
\qquad \qquad
f_{8/9}(\mathbf{x})=d_F(\mathbf{x}) \pm d_P(\mathbf{x}).
\vspace{-2mm}
\end{equation}
These features are useful because the sum features $f_6$ and $f_8$ measure whether voxel $\mathbf{x}$ locates within the narrow space between two bones, and the difference features $f_7$ and $f_9$ measure which bone it is closer to. Fig. \ref{fig:feature_scatter} shows how $f_6$ and $f_7$ in addition to intensity feature $f_1$ separate tibial cartilage from femoral and patellar cartilages.

Given the prior knowledge that the cartilage can only grow in certain area on the bone surface, it is useful for the cartilage segmentation to not only know how close the voxel is to the bone surface, but also where it is anatomically. Therefore we define the distance features to the densely registered landmarks on the bone surface as described in Section \ref{sec:2}:
$f_{10}(\mathbf{x,\zeta})=|| \mathbf{x}-\mathbf{z}_\zeta ||$,
where $\mathbf{z}_\zeta$ is the spatial coordinate of the $\zeta$th landmark of all bone mesh points. $\zeta$ is randomly generated in training due to the great number of mesh points available (Fig. \ref{fig:d2lm}).

\vspace{-4mm}
\begin{table}
\begin{center}
\begin{tabular}{c|c|c|c}
\hline
 &
\multicolumn{1}{c|}{Femur DSC (\%)}  &
\multicolumn{1}{c|}{Tibia DSC (\%)} &
\multicolumn{1}{c}{Patella DSC (\%)}
\\
\hline
Before RW & $92.37 \pm 1.58$ & $94.64  \pm1.18$ & $92.07 \pm 1.47$ \\
After RW & $94.86 \pm 1.85$ & $95.96 \pm 1.64$ & $94.31 \pm 2.15$
\\ \hline
\end{tabular}
\end{center}
\caption{
The Dice similarity coefficient (DSC) of bone segmentation results before and after random walks (3-fold cross validation on 176 OAI volumes). }
\label{tab:dsc_bone}
\end{table}

\begin{figure}[h!]
  \begin{minipage}{0.47\textwidth}
    \subfloat[]
    {\label{fig:bone_seg}\includegraphics[width=0.95\textwidth]{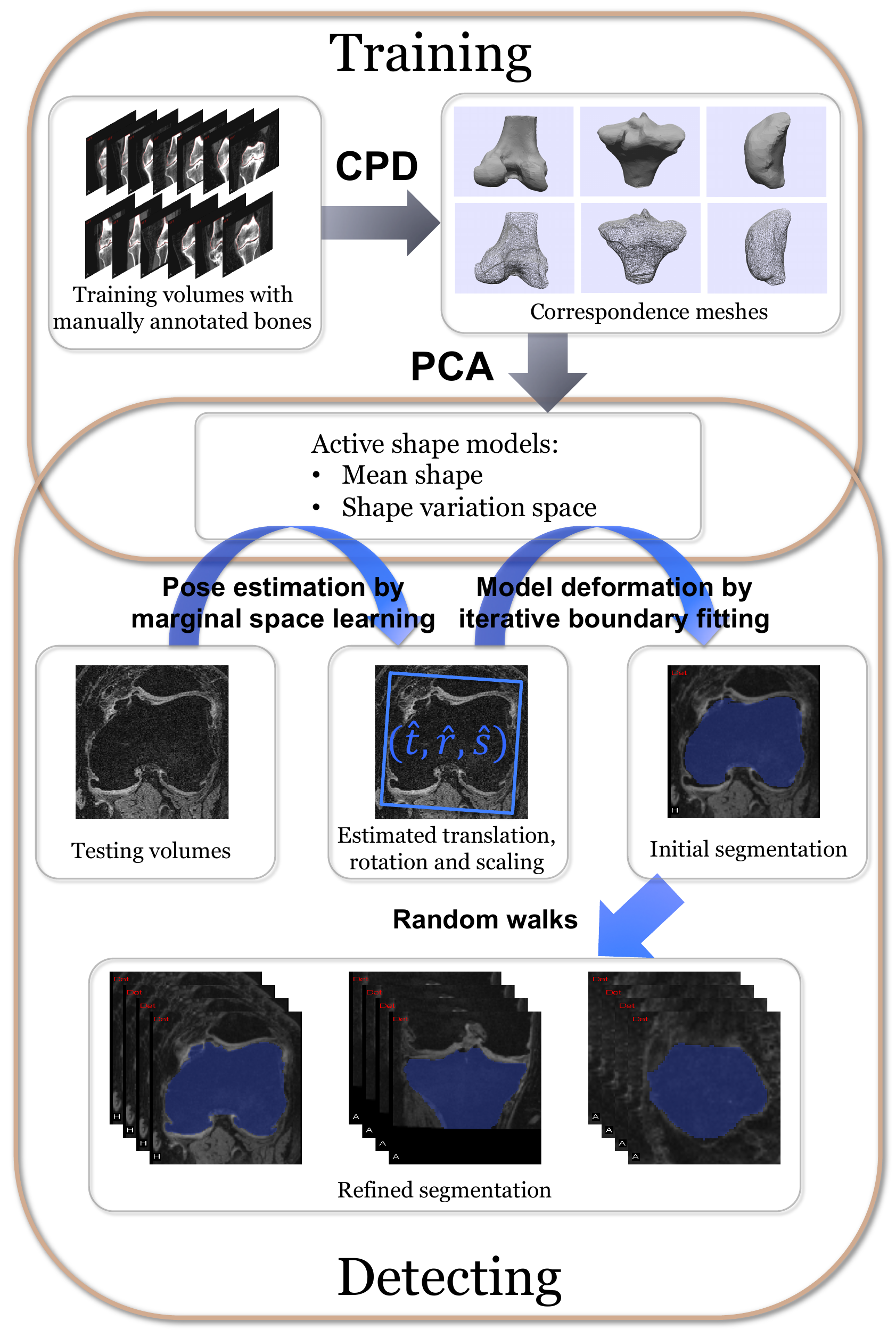}}
  \end{minipage}
  \hfill
  \begin{minipage}{0.47\textwidth}
    \subfloat[]
    {\label{fig:anatomy1}\includegraphics[height=0.43\textwidth]{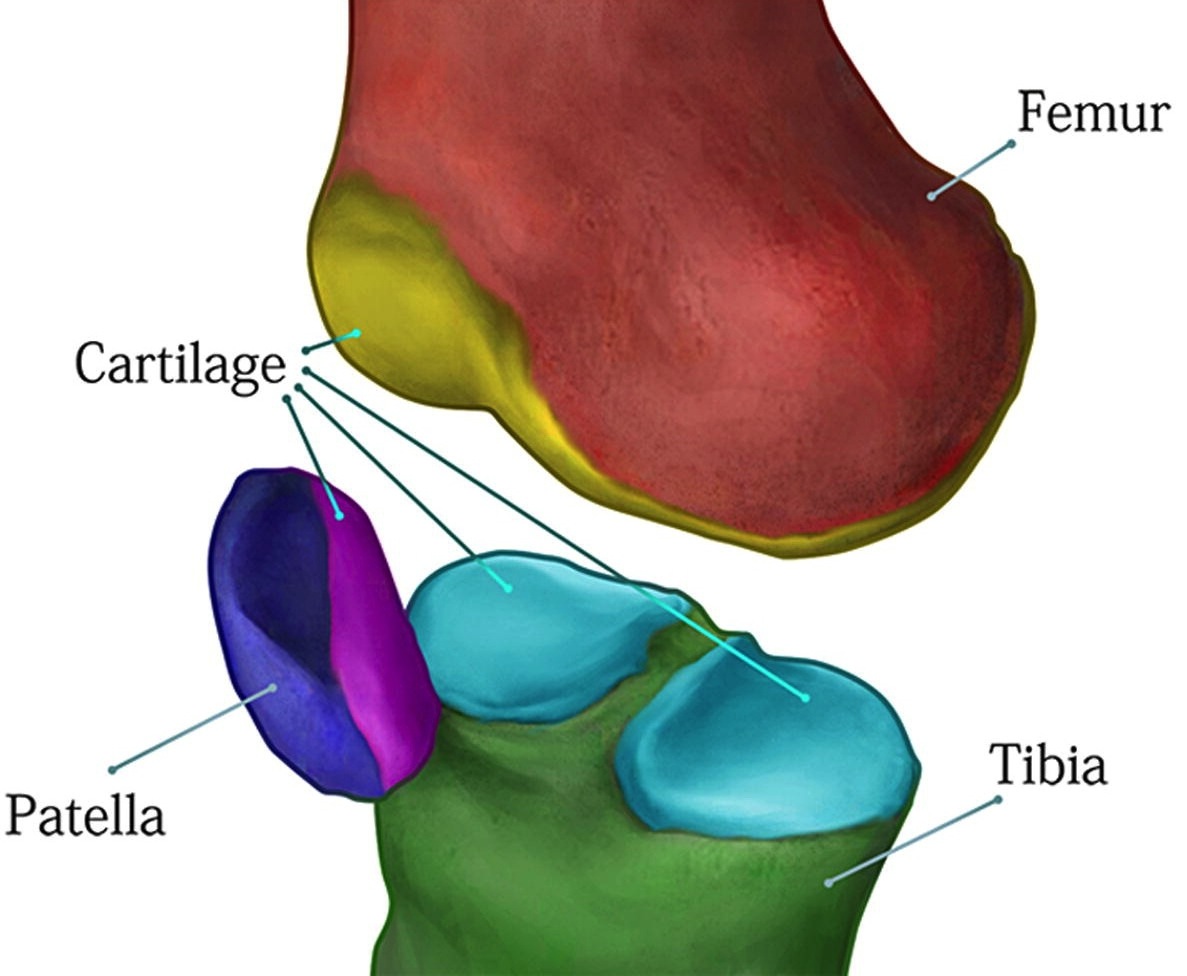}}
    \quad
    \subfloat[]
    {\label{fig:anatomy2}\includegraphics[height=0.43\textwidth]{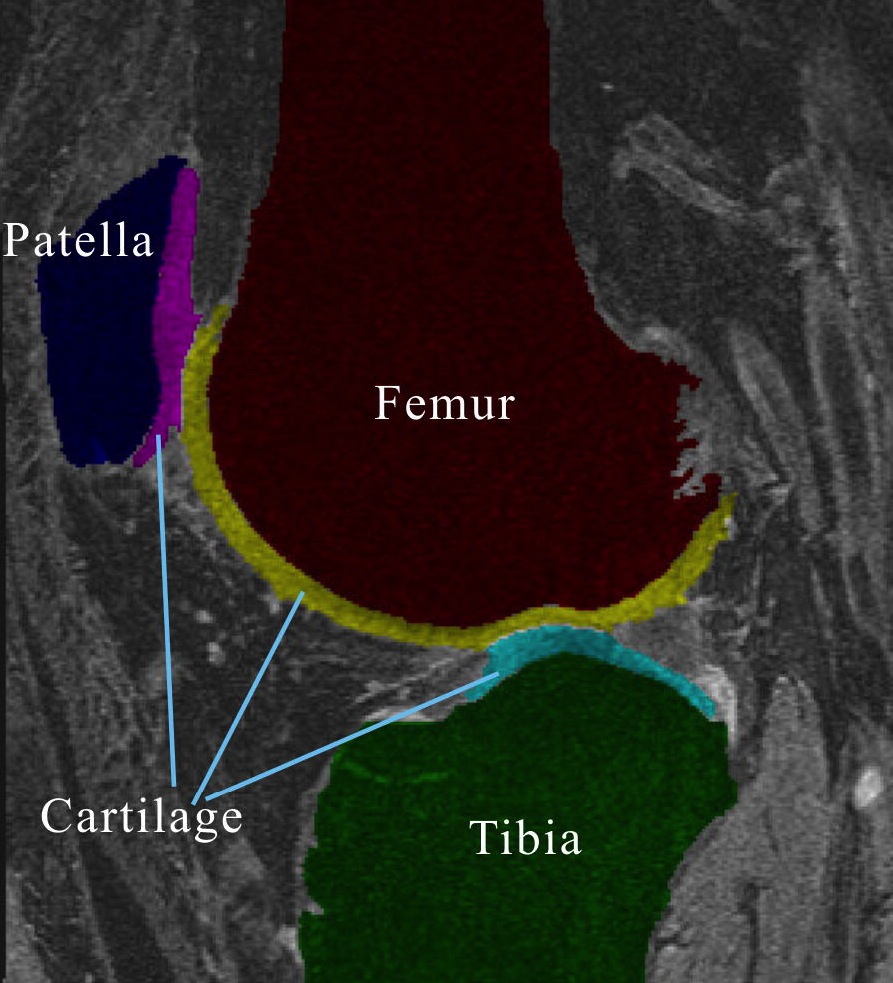}}
    \\
    \subfloat[]
    { \label{fig:2_pass_forest}\includegraphics[width=0.95\textwidth]{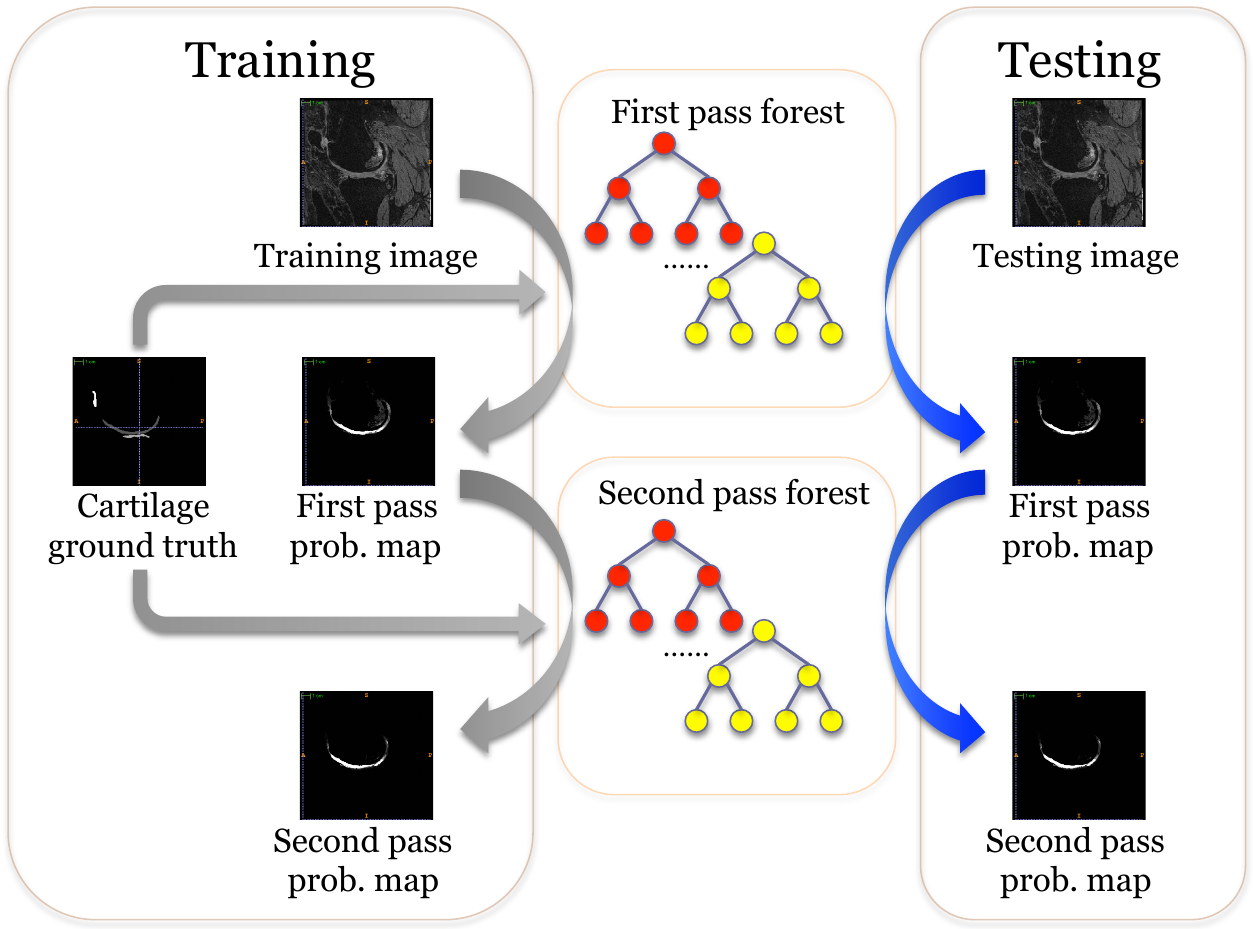}}
  \end{minipage}
  \caption{
  (a) The bone segmentation framework.
  (b) 3D anatomy of knee joint.
  (c) Example of a 2D MR slice \cite{opt_prob}.
  (d) The semantic context forests diagram.}

\end{figure}

\vspace{-6mm}

\begin{figure*}[h!]
  \centering
    \subfloat[]
      {\label{fig:prob_0}\includegraphics[width=0.23\textwidth]{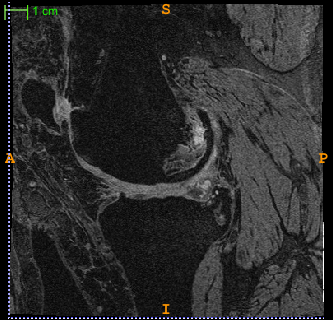}} \,
    \subfloat[]
      {\label{fig:prob_1}\includegraphics[width=0.23\textwidth]{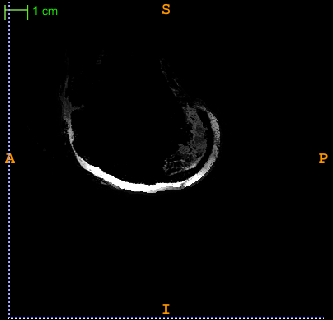}} \,
    \subfloat[]
      {\label{fig:prob_2}\includegraphics[width=0.23\textwidth]{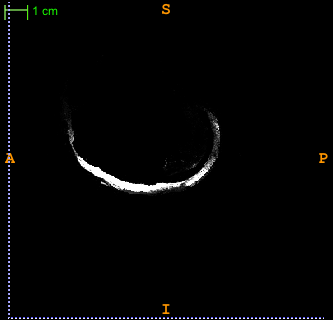}} \,
    \subfloat[]
      {\label{fig:prob_2}\includegraphics[width=0.23\textwidth]{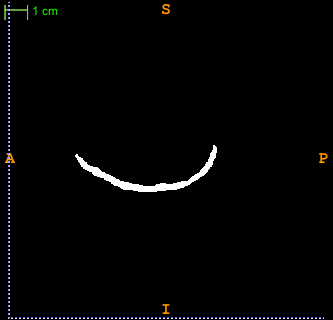}}
  \caption{
  Probability maps of femoral cartilage by semantic context forests.
  (a) Original image. (b) Prob. map of the 1st pass.
  (c) Prob. map of the 2nd pass.
  (d) Ground truth.          }
  \label{fig:prob_map}
\end{figure*}

\vspace{-4mm}

\textbf{Context Features} compare the intensity of the current voxel $\mathbf{x}$ and another voxel $\mathbf{x}+\mathbf{u}$ with random offset $\mathbf{u}$:
$f_{11}(\mathbf{x,u})=I(\mathbf{x+u})-I(\mathbf{x})$,
where $\mathbf{u}$ is a random offset vector. This subset of features, named as random shift intensity difference (RSID) features in this paper, capture the context information in different ranges by randomly generating a large number of different values of $\mathbf{u}$ from a uniform distribution in training. They were earlier used to solve pose classification \cite{kinect} and keypoint recognition  \cite{keypoint_recognition} problems.

\subsection{Iterative Semantic Context Forests}

In this paper, we present a multi-pass iterative classification method to automatically exploit the semantic context for multiple object segmentation problems. In each pass, the generated probability maps will be used to extract the context embedded features to enhance the classification performance of the next pass. Fig. \ref{fig:2_pass_forest} shows a 2-pass iterative classification framework with the random forests  \cite{kinect,keypoint_recognition,decision_tree_1986,random_forest_2001,tracking_trees,zikic} selected as the base classifier for each pass.
However, the method can be extended to more iterations with the use of other discriminative classifiers.

\subsubsection*{Semantic Context Features}
After each pass of the classification, the probability maps are generated and used to extract semantic context features as defined below:
$f_{12}(\mathbf{x})=P_F(\mathbf{x})$, $f_{13}(\mathbf{x})=P_T(\mathbf{x})$, $f_{14}(\mathbf{x})=P_P(\mathbf{x})$,
where $P_F$, $P_T$ and $P_P$ stand for the femoral, tibia and patellar cartilage probability map, respectively. In the same fashion as the RSID features, we compare the probability response of two voxels with random shift:
\vspace{-2mm}
\begin{eqnarray}
f_{15/16/17}(\mathbf{x},\mathbf{u})&=&P_{F/T/P}(\mathbf{x+u})-P_{F/T/P}(\mathbf{x}),
\vspace{-2mm}
\end{eqnarray}
which is called random shift probability difference features (RSPD). RSPD provides semantic context information because the probability map values are directly associated with anatomical labels, rather than original intensity volume.

In Fig. \ref{fig:prob_map}, it can be observed that the probability map of the second pass classification is significantly enhanced with much less noisy responses, compared with the first pass.

\subsection{Post-processing by Graph Cuts Optimization}

After the classification, we finally use the probabilities of being the background and the three cartilages to construct the energy functions and perform multi-label graph cuts \cite{GCO} to refine the segmentation with smoothness constraints.

The graph cuts algorithm assigns a label $l(\mathbf{x})$ to each voxel $\mathbf{x}$, such that the energy below is minimized:
\begin{eqnarray}
E(L)=\sum\limits_{\lbrace \mathbf{x,y} \rbrace \in \mathcal{N} } V_{\mathbf{x,y}}(l(\mathbf{x}),l(\mathbf{y}))
+\sum\limits_{\mathbf{x}} D_{\mathbf{x}}(l(\mathbf{x})),
\end{eqnarray}
where $L$ is the global label configuration, $\mathcal{N}$ is the neighborhood system, $V_{\mathbf{x,y}}(\cdot)$ is the smoothness energy, and $D_{\mathbf{x}}(\cdot)$ is the data energy.
We define
\begin{eqnarray}
D_{\mathbf{x}}(l(\mathbf{x}))&=&-\lambda \ln P_{l(\mathbf{x})}(\mathbf{x}), \\
V_{\mathbf{x,y}}(l(\mathbf{x}),l(\mathbf{y}))&=&\delta_{l(\mathbf{x}) \neq l(\mathbf{y})}e^{\frac{(I(\mathbf{x})-I(\mathbf{y}))^2}{2\sigma^2}}.
\end{eqnarray}
$\delta_{l(\mathbf{x}) \neq l(\mathbf{y})}$ takes value 1 when $l(\mathbf{x})$ and $l(\mathbf{y})$ are different labels, and takes value 0 when $l(\mathbf{x})=l(\mathbf{y})$. $P_{l(\mathbf{x})}(\mathbf{x})$ takes the value $P_{F}(\mathbf{x})$, $P_{T}(\mathbf{x})$, $P_{P}(\mathbf{x})$ or $1-P_{F}(\mathbf{x})-P_{T}(\mathbf{x})-P_{P}(\mathbf{x})$, depending on the label $l(\mathbf{x})$. $\lambda$ and $\sigma$ are two parameters. $\lambda$ specifies the weight of data energy versus smoothness energy, while $\sigma$ is associated with the image noise \cite{graph_cuts}.

\vspace{-2mm}
\section{Experimental Results}
\vspace{-2mm}
\label{sec:4}
\subsection{Dataset and Experiment Settings}

The dataset we use in our work is the publicly available Osteoarthritis Initiative (OAI) dataset, which contains both 3D MR images and ground truth cartilage annotations, referred to as ``kMRI segmentations (iMorphics)''. The sagittal 3D 3T (Tesla) DESS (dual echo steady state) WE (water-excitation) MR images in OAI have high-resolution, good delineation of articular cartilage, fast acquisition time and high SNR. Our dataset consists of 176 volumes from 88 subjects, and belongs to the Progression subcohort, where all subjects show symptoms of OA. Each subject has two volumes scanned in different years. The size of each image volume is $384 \times 384 \times 160$ voxels, and the voxel size is $0.365 \times 0.365 \times 0.7 \, \mathrm{mm^3}$.

For the validation, we divide the OAI dataset to three equally-sized subsets: $D_1$, $D_2$ and $D_3$, and perform a
three-fold validation. The two volumes from the same subject are always placed in the same subset. For each randomized decision tree, we set the depth of the tree to 18, and train 60 trees in each pass. During training, the number of candidates at each non-leaf node is set to 1000. The dice similarity coefficient (DSC) is used to measure the performance of our method since it is commonly reported in previous literature \cite{voxel_classification,fripp,logismos,opt_prob,shan_liang}.

\subsection{Results}

First, we compare the frequency of different features that is selected by the classifiers. As shown in Fig. \ref{fig:feature_frequency}, RSID, RSPD and the distance to dense landmarks are very informative features to embed spatial constraints.

Then we compare the segmentation performance with and without the use of the distance features to the anatomical dense  landmarks, and also the results with different number of classification iterations. The results in Fig. \ref{fig:bar} demonstrate the effectiveness of the distance features to dense landmarks and iterative classification with semantic context forests. In particular, 2-pass random forests achieve significant performance improvement, whereas the gain seems quite negligible by adding more passes.

\begin{figure}
  \centering
  \includegraphics[width=0.95\textwidth]{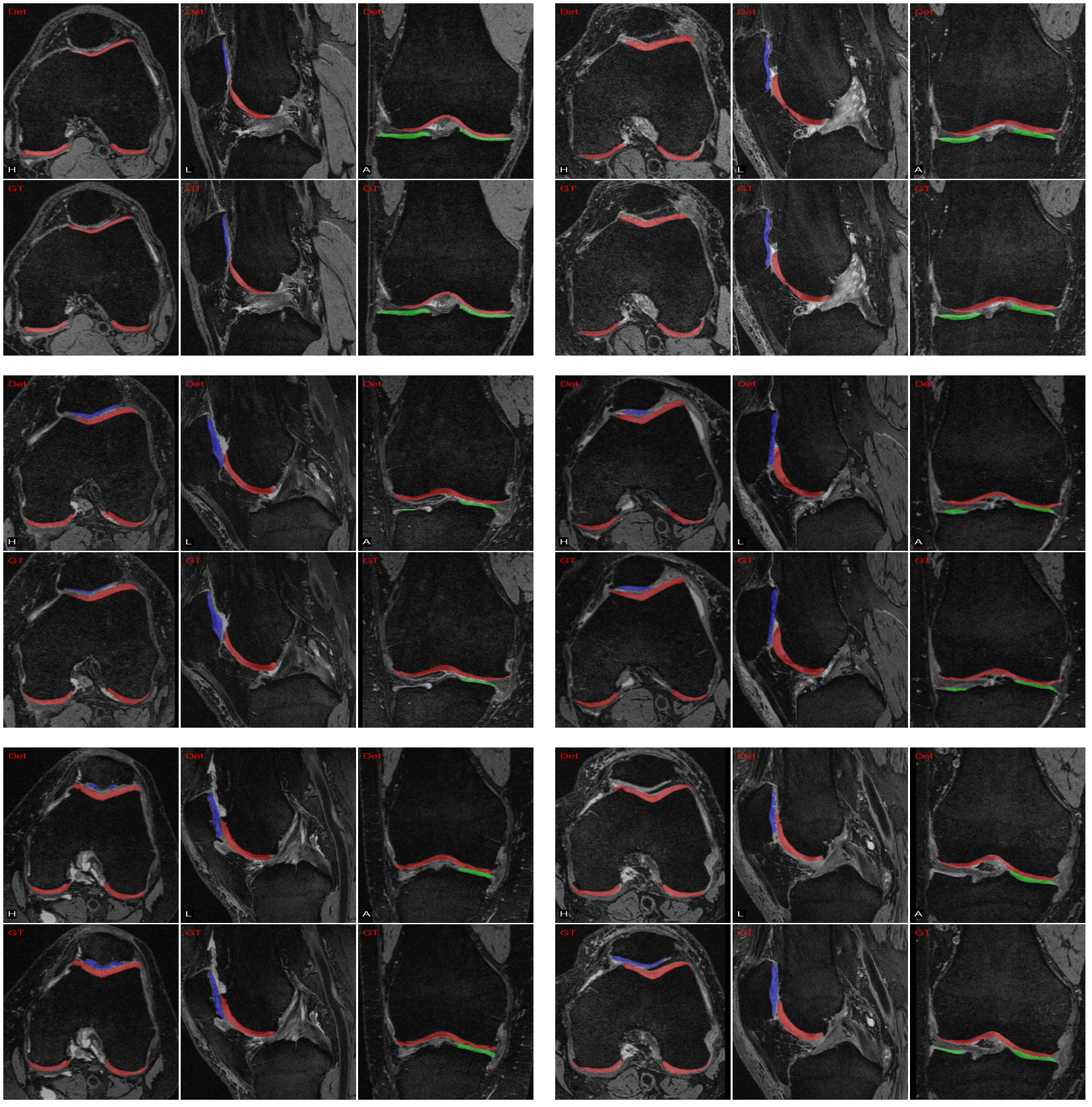}
  \caption{Examples cartilage segmentations compared with ground truth.
  Each case has six images: segmentation results in upper row, ground truth in lower row.
  Red: femoral cartilage; green: tibial cartilage; blue: patellar cartilage. }
  \label{fig:many_result}
  \vspace{-3mm}
\end{figure}

Finally, the quantitative results (2-pass classification) are listed in Table \ref{tab:dsc} together with the numbers reported in the earlier literature. Because the datasets used are different by all these approaches, the numbers in the table are only for reference. Note that only our experiments are based on a relatively large dataset. As shown in the table, we achieved high performance with regard to the femoral and tibial cartilage, whereas the DSC of patellar cartilage is notably lower than the other two cartilages. This is partly because the size of patellar cartilage is much smaller than femoral and tibial cartilage, so that the same amount of segmentation error will result in lower DSC. Besides, some patellar cartilage annotations in the dataset do not appear very consistent with others. Example segmentation results are shown in
Fig. \ref{fig:many_result}.

\vspace{-2mm}
\section{Conclusion}
\vspace{-2mm}
\label{sec:5}
We have presented a new approach to segment the three knee cartilages in 3-D MR images, which effectively exploits the semantic context information in the knee joint. By using the distance features to the bone surface as well as to the dense anatomical landmarks on the bone surface, the spatial constraints between cartilages and bones are incorporated without the need of explicit extraction of the bone cartilage interface. Furthermore, the use of multi-pass iterative classification with semantic context forests provides more spatial constraints between different cartilages to further improve the segmentation. The experiment validation shows the effectiveness of this method. Ongoing work include the joint bone-and-cartilage voxel classification  in the iterative classification framework.

\begin{table*}
\scriptsize
\begin{center}
\begin{tabular}{c|c|cc|cc|cc}
\hline
 & &
\multicolumn{2}{c|}{Fem. Cart. DSC}  &
\multicolumn{2}{c|}{Tib. Cart. DSC} &
\multicolumn{2}{c}{Pat. Cart. DSC}
\\
Author  & Dataset &
Mean & Std. &
Mean & Std. &
Mean & Std. \\
\hline
Shan \cite{shan_liang} & 18 SPGR images & 78.2\% & 5.2\% & 82.6\% & 3.8\% & -- & -- \\
Folkesson \cite{voxel_classification} & 139 Esaote C-Span images & 77\% & 8.0\% & 81\% & 6.0\% & -- & -- \\
Fripp \cite{fripp} & 20 FS SPGR images & 84.8\% & 7.6\% & 82.6\% & 8.3\% & 83.3\% & 13.5\% \\
Lee \cite{opt_prob} & 10 images in OAI & 82.5\% & -- & 80.8\% & -- & 82.1\% & -- \\
Yin \cite{logismos} & 60 images in OAI & 84\% & 4\% & 80\% & 4\% & 80\% & 4\% \\
\hline

 & OAI, $D_1$ subset (58 images) &  85.47\% & 3.10\% & 84.96\% & 3.82\% & 78.56\% & 9.38\%  \\
Proposed & OAI, $D_2$ subset (58 images) &  85.20\% & 3.65\% & 83.52\% & 4.08\% & 80.79\% & 7.40\%  \\
method & OAI, $D_3$ subset (60 images) &  84.22\% & 3.05\% & 82.74\% & 3.84\% & 78.12\% & 9.63\%  \\
 & OAI, overall (176 images)  &  84.96\% & 3.30\% & 83.74\% & 4.00\% & 79.16\% & 8.88\%  \\
\hline
\end{tabular}
\end{center}

\caption{
Performance of our method compared with other state-of-the-art cartilage segmentation methods: mean DSC and standard deviation.}
\label{tab:dsc}

\end{table*}

\vspace{-2mm}

{
\bibliographystyle{splncs}
\bibliography{egbib}

@article{QAOC-2004,
  author  = {Graichen, H. and Eisenhart-Rothe, R. and Vogl, T. and Englmeier, K. H. and Eckstein, F.},
  title   = {Quantitative assessment of cartilage status in osteoarthritis by quantitative magnetic resonance imaging},
  journal = {Arthritis Rheumatism},
  year    = {2004}
}

@article{voxel_classification,
  author  = {Folkesson, J. and Dam, E. and Olsen, O. and Pettersen, P. and Christiansen, C.},
  title   = {Segmenting articular cartilage automatically using a voxel classification approach},
  journal = {IEEE Trans. Med. Imag.},
  volume  = {26},
  number  = {1},
  month   = {Jan.},
  year    = {2007},
  pages   = {106--115}
}

@inproceedings{FASO-2010,
  author    = {Vincent, G. and Wolstenholme, C. and Scott, I. and Bowes, M.},
  title     = {Fully automatic segmentation of the knee joint using active appearance models},
  booktitle = {Medical Image Analysis for the Clinic: A Grand Challenge},
  year      = {2010}
}

@article{fripp,
  author  = {Fripp, J. and Crozier, S. and Warfield, S. and Ourselin, S.},
  title   = {Automatic segmentation and quantitative analysis of the articular cartilages from magnetic resonance images of the knee},
  journal = {IEEE Trans. Med. Imag.},
  volume  = {29},
  number  = {1},
  month   = {Jan.},
  year    = {2010},
  pages   = {55--64}
}

@article{logismos,
  author  = {Yin, Y. and Zhang, X. and Williams, R. and Wu, X. and Anderson, D. and Sonka, M.},
  title   = {Logismos -- layered optimal graph image segmentation of multiple objects and surfaces: Cartilage segmentation in the knee joint},
  journal = {IEEE Trans. Med. Imag.},
  volume  = {29},
  number  = {12},
  month   = {Dec.},
  year    = {2010},
  pages   = {2023--2037}
}

@article{opt_prob,
  author  = {Lee, S. and Park, S. H. and Shim, H. and Yun, I. D. and Lee, S. U.},
  title   = {Optimization of local shape and appearance probabilities for segmentation of knee cartilage in 3-D MR images},
  journal = {CVIU},
  volume  = {115},
  number  = {12},
  month   = {Dec.},
  year    = {2011},
  pages   = {1710--1720}
}

@article{multicolumn_graph_cut,
  author  = {Li, K. and Wu, X. and Chen, D. and Sonka, M.},
  title   = {Optimal surface segmentation in volumetric images--a graph-theoretic approach},
  journal = {IEEE Trans. PAMI},
  volume  = {28},
  number  = {1},
  month   = {Jan.},
  year    = {2006},
  pages   = {119--134}
}

@article{auto-context,
  author  = {Tu, Z. and Bai, X.},
  title   = {Auto-context and its application to high-level vision tasks and 3D brain image segmentation},
  journal = {IEEE Trans. PAMI},
  month   = {Oct.},
  year    = {2010},
  pages   = {1744--1757}
}

@inproceedings{ipmi2011,
  author    = {Montillo, A. and Shotton, J. and Winn, J. and Iglesias, J. and Metaxas, D. and Criminisi, A.},
  title     = {Entangled decision forests and their application for semantic segmentation of CT images},
  booktitle = {IPMI},
  year      = {2011},
  pages     = {184--196}
}

@inproceedings{HLBA-2008,
  author    = {Ling, H. and Zheng, Y. and Georgescu, B. and Zhou, S. K. and Suehling, M.},
  title     = {Hierarchical learning-based automatic liver segmentation},
  booktitle = {CVPR},
  year      = {2008}
}

@article{FCHM-2008,
  author  = {Zheng, Y. and Barbu, A. and Georgescu, M. and Scheuring, M. and Comaniciu, D.},
  title   = {Four-chamber heart modeling and automatic segmentation for 3{D} cardiac {CT} volumes using marginal space learning and steerable features},
  journal = {IEEE Trans. Med. Imag.},
  volume  = {27},
  number  = {11},
  year    = {2008},
  pages   = {1668--1681}
}

@article{CPD,
  author  = {Myronenko, A. and Song, X.},
  title   = {Point set registration: Coherent point drift},
  journal = {IEEE Trans. PAMI},
  volume  = {32},
  number  = {12},
  month   = {Dec.},
  year    = {2010},
  pages   = {2262--2275}
}

@article{ASM,
  author  = {Cootes, T. and Taylor, C. and Cooper, D. and Graham, J.},
  title   = {Active shape models--their training and application},
  journal = {CVIU},
  volume  = {61},
  number  = {1},
  year    = {1995},
  pages   = {38--59}
}

@article{random_walk,
  author  = {Grady, L.},
  title   = {Random walks for image segmentation},
  journal = {IEEE Trans. PAMI},
  volume  = {28},
  number  = {11},
  month   = {Nov.},
  year    = {2006},
  pages   = {1768--1783}
}

@inproceedings{kinect,
  author    = {Shotton, J. and Fitzgibbon, A. and Cook, M. and Sharp, T. and Finocchio, M. and Moore, R. and Kipman, A. and Blake, A.},
  title     = {Real-time human pose recognition in parts from single depth images},
  booktitle = {CVPR},
  month     = {Jun.},
  year      = {2011},
  pages     = {1297--1304}
}

@inproceedings{keypoint_recognition,
  author    = {Lepetit, V. and Lagger, P. and Fua, P.},
  title     = {Randomized trees for real-time keypoint recognition},
  booktitle = {CVPR},
  volume    = {2},
  month     = {Jun.},
  year      = {2005},
  pages     = {775--781 vol. 2}
}

@article{decision_tree_1986,
  author  = {Quinlan, J. R.},
  title   = {Induction of decision trees},
  journal = {Machine learning},
  volume  = {1},
  number  = {1},
  year    = {1986},
  pages   = {81--106}
}

@article{random_forest_2001,
  author  = {Breiman, L.},
  title   = {Random forests},
  journal = {Machine learning},
  volume  = {45},
  number  = {1},
  year    = {2001},
  pages   = {5--32}
}

@inproceedings{tracking_trees,
  author    = {Wang, Q. and Ou, Y. and Julius, A. A. and Boyer, K. L. and Kim, M. J.},
  title     = {{Tracking \textit{Tetrahymena Pyriformis} Cells using Decision Trees}},
  booktitle = {ICPR},
  month     = {Nov.},
  year      = {2012}
}

@misc{zikic,
  author = {Zikic, D. and Glocker, B. and Konukoglu, E. and Shotton, J. and Criminisi, A. and Ye, D. and Demiralp, C. and Thomas, O. and Das, T. and Jena, R. and others},
  title  = {Context-sensitive classification forests for segmentation of brain tumor tissues, {MICCAI}},
  year   = {2012}
}

@article{GCO,
  author  = {Boykov, Y. and Veksler, O. and Zabih, R.},
  title   = {Fast approximate energy minimization via graph cuts},
  journal = {IEEE Trans. PAMI},
  volume  = {23},
  number  = {11},
  month   = {Nov.},
  year    = {2001},
  pages   = {1222--1239}
}

@article{graph_cuts,
  author  = {Boykov, Y. and Funka-Lea, G.},
  title   = {Graph cuts and efficient N-D image segmentation},
  journal = {Int. J. Comput. Vis.},
  volume  = {70},
  number  = {2},
  month   = {Nov.},
  year    = {2006},
  pages   = {109--131}
}

@inproceedings{shan_liang,
  author    = {Shan, L. and Charles, C. and Niethammer, M.},
  title     = {Automatic atlas-based three-label cartilage segmentation from MR knee images},
  booktitle = {2012 IEEE Workshop on Mathematical Methods in Biomedical Image Analysis},
  month     = {Jan.},
  year      = {2012},
  pages     = {241--246}
}
}

\clearpage
\appendix
\renewcommand{\theHsection}{\Alph{section}}
\emergencystretch=2em

\section{Open-Source Implementation (\texttt{kneeseg})}
\label{sec:appendix_code}

To facilitate reproducible research and provide an accessible, self-contained benchmark for learning-based 3D knee MRI bone and cartilage segmentation, we have released a modern, pure-Python open-source implementation of this work: \texttt{kneeseg} (version 0.2.5). Table~\ref{tab:opensource_links} summarizes the official open-source code repository, Python package distribution, and pretrained model collection.

\begin{table}[h!]
\centering
\scriptsize
\setlength{\tabcolsep}{5pt}
\renewcommand{\arraystretch}{1.25}
\caption{Summary of open-source resources for \texttt{kneeseg}.}
\label{tab:opensource_links}
\begin{tabularx}{\textwidth}{l X}
\toprule
\textbf{Resource} & \textbf{Location / Details} \\
\midrule
GitHub Repository & \url{https://github.com/wq2012/kneeseg} \\
PyPI Package & \texttt{kneeseg} (v0.2.5) --- \url{https://pypi.org/project/kneeseg/} \\
HF Model Collection & \url{https://huggingface.co/collections/wq2012/knee-mri-segmentation} \\
License & MIT License (source code and all five Hugging Face pretrained models) \\
Core Dependencies & \texttt{numpy}, \texttt{scipy}, \texttt{scikit-learn}, \texttt{scikit-image}, \texttt{SimpleITK}, \texttt{ml\_dtypes}, \texttt{pycpd}, \texttt{joblib}, \texttt{threadpoolctl}, \texttt{tqdm}, \texttt{jsonschema} \\
\bottomrule
\end{tabularx}
\end{table}

\subsection{Algorithmic Evolution and Differences from the Original Paper}
\label{sec:appendix_diff}

While the original 2013 paper relied on proprietary internal C++ libraries at Siemens Corporate Research---specifically Marginal Space Learning (MSL)~\cite{FCHM-2008} and Active Shape Model (ASM) mesh deformation~\cite{ASM} for pre-segmenting the bones---the open-source \texttt{kneeseg} library provides a completely self-contained, end-to-end 3D segmentation pipeline using open-source scientific Python libraries. Furthermore, to achieve high accuracy across both high-resolution anisotropic clinical scans (SKI10) and lower-resolution isotropic scans (downsampled OAI) without requiring ground-truth bone meshes at inference time, \texttt{kneeseg} introduces several key algorithmic enhancements over the original formulation. Table~\ref{tab:algo_comparison} summarizes the component-by-component differences, which are detailed below.

\begin{table}[t!]
\centering
\scriptsize
\setlength{\tabcolsep}{4pt}
\renewcommand{\arraystretch}{1.22}
\caption{Comparison between the original 2013 paper implementation and the open-source \texttt{kneeseg} (v0.2.5) reproduction.}
\label{tab:algo_comparison}
\begin{tabularx}{\textwidth}{p{2.3cm} X X}
\toprule
\textbf{Pipeline Stage} & \textbf{Original 2013 Paper (Siemens C++)} & \textbf{Open-Source \texttt{kneeseg} (Python v0.2.5)} \\
\midrule
\textbf{Bone Segmentation} (Section~\ref{sec:2}) &
3-step mesh pipeline: Marginal Space Learning (MSL) pose estimation $(\hat{t},\hat{r},\hat{s})$, Active Shape Model (ASM) mesh deformation with Coherent Point Drift (CPD) correspondences, and 3D Random Walks (RW) refinement. &
Self-contained 2-Pass Physical-Space Auto-Context Forest (\texttt{bone\_rf\_p1}, \texttt{bone\_rf\_p2}) + Binary Subchondral Articular Surface Refiner (\texttt{RefBone V2}, blended with \texttt{RefBone V1} on SKI10) with 1D cortical enclosure rays, 3D bone atlases, Patellofemoral Specialist, and shaft hole-filling cleanup. Standalone CPD and 3D RW modules also included. \\
\midrule
\textbf{Bone Distance Features} ($f_3$--$f_9$, Section~\ref{sec:3}) &
Signed Euclidean distances $d_F, d_T, d_P$ to refined bone meshes and sum/difference combinations $d_F \pm d_T$, $d_F \pm d_P$. &
Exact physical-millimeter Signed Maurer Distance Transforms (\texttt{SimpleITK}) $d_F, d_T, d_P$ and linear combinations $d_F \pm d_T$, $d_F \pm d_P$ computed directly from predicted 3D bone masks. \\
\midrule
\textbf{Anatomical Localization} ($f_{10}$, Section~\ref{sec:3}) &
Euclidean distances $\|\mathbf{x}-\mathbf{z}_\zeta\|$ to densely registered CPD bone mesh points $\mathbf{z}_\zeta$. &
Deterministic quantile-based 18 articular surface landmarks per bone ($3\times 3$ condylar/plateau grids) + multi-resolution 4D Articular Coordinate Probabilistic Atlases (8 tables on SKI10, 11 on OAI) built via Leave-One-Out (LOO) cross-validation. \\
\midrule
\textbf{Appearance \& Context Features} ($f_1, f_2, f_{11}$) &
Raw intensity $I(\mathbf{x})$, gradient magnitude $\|\nabla I(\mathbf{x})\|$, and voxel-offset Random Shift Intensity Differences (RSID) $I(\mathbf{x}+\mathbf{u})-I(\mathbf{x})$. &
ROI interquartile ($p_{10}, p_{50}, p_{90}$) normalized intensity, multi-scale physical-mm Gaussian, gradient, Laplacian, and structure-tensor features, physical-mm RSID, and 1D surface-normal ray profiles ($s \in [-3.6, +5.6]\,\mathrm{mm}$) along $\nabla d_{\mathrm{bone}}$. \\
\midrule
\textbf{Iterative Semantic Context Forests} ($f_{12}$--$f_{17}$, Section~\ref{sec:3}) &
2-pass custom C++ Random Forest (60 trees, depth 18, 1000 node candidates). Pass 2 extracts class probabilities $P_F, P_T, P_P$ and RSPD features from in-sample Pass-1 predictions. &
4-Class Compartment-Specialist Ensemble (\texttt{ExtraTrees} + \texttt{HistGradientBoosting}) predicting \{\texttt{Background}, \texttt{Subchondral Bone/Meniscus}, \texttt{Target Cart.}, \texttt{Opposing Cart.}\}. Pass-2 RSPD and normal-ray probability profiles are trained via 4-fold Out-Of-Fold (OOF) stacking. \\
\midrule
\textbf{Post-Processing} (Section~\ref{sec:3}) &
Multi-label Graph Cuts energy minimization $E(L) = \sum V_{\mathbf{x},\mathbf{y}} + \sum D_{\mathbf{x}}$. &
4-fold OOF Stage-3 Subchondral Boundary Refiner (multi-family 103--167 features on SKI10; 101-feature specialist on OAI) + OOF Ridge Threshold \& Bone-Penalty Calibrator $(\mathrm{th}, \mathrm{bp})$ + physical-space Gaussian smoothing + contact resolution and component filtering. \\
\midrule
\textbf{Cross-Domain Support} &
Single-domain training and evaluation on full-resolution OAI ($0.365 \times 0.365 \times 0.7\,\mathrm{mm}^3$). &
Automatic C/Fortran memory-layout and canonical $(Z,Y,X)$ orientation detection; geometry-aware routing + 208-scan cross-domain atlas regularization across SKI10 and OAI. \\
\bottomrule
\end{tabularx}
\end{table}

\subsubsection{1. Self-Contained 2-Pass Bone Forest and Subchondral Surface Refiner.}
Because the original MSL pose detectors and proprietary statistical shape models from Section~\ref{sec:2} are not publicly redistributable, \texttt{kneeseg} replaces mesh-based bone pre-segmentation with a 2-pass physical-space voxel and boundary-shell classification cascade (\nolinkurl{kneeseg/bone_rf.py} and \nolinkurl{kneeseg/oai_subchondral_atlas.py}):
\begin{itemize}
    \item \textbf{Pass 1 (\texttt{bone\_rf\_p1.joblib}):} A joint-anchored multi-scale physical-space ensemble (\texttt{ExtraTreesClassifier} and \texttt{HistGradientBoostingClassifier}) operating on a strided physical grid ($1\times 2\times 2$ voxel stride). It extracts percentile-normalized MRI intensities ($p_5, p_{50}, p_{95}$), multi-scale physical-millimeter Gaussian-smoothed intensities ($\sigma \in \{1, 2, 4, 8, 14, 24\}\,\mathrm{mm}$) and gradient magnitudes, local z-scores, slice-wise axial normalizations, 1D directional minimum/maximum cortical enclosure ray filters ($18\,\mathrm{mm}$ and $45\,\mathrm{mm}$ rays along $\pm Z, \pm Y, \pm X$ capturing the dark cortical shell enclosing bright cancellous bone), joint-anchored 3D coordinate frames $(c_j, \mathrm{span}_z)$, 3D probabilistic bone atlases, and physical-millimeter RSID offsets (139/142 features on SKI10; 130 Pass-0 and 151 Tier-A features on OAI). On 6-class OAI scans, Pass~1 is augmented by a dedicated 109-feature Patellofemoral Specialist (\nolinkurl{extract_patella_specialist_features}) anchored on the anterior femur. Predictions undergo 3D topological post-processing (\nolinkurl{postprocess_bone_with_shaft_fill} and \nolinkurl{postprocess_3bone_mask}) that retains the largest connected component per bone and fills internal medullary shaft cavities slice-by-slice.
    \item \textbf{Pass 2 (\texttt{bone\_rf\_p2.joblib}):} While the baseline \nolinkurl{BoneClassifier.train()} API supports a 4-class boundary-band formulation (\texttt{Background}, \texttt{Bone}, \texttt{Articular Cartilage}, \texttt{Meniscus/Soft Tissue}), the released Subchondral Atlas checkpoints use a full-resolution binary boundary-shell auto-context forest and Subchondral Articular Surface Refiner (\texttt{RefBone V2}, $\texttt{classes\_} = [0, 1]$ for \texttt{Non-Bone} vs.\ \texttt{Target Bone}). Operating within the subchondral boundary shell ($|d_{\mathrm{bone}}| \le 3.0\,\mathrm{mm}$ on SKI10; $|d_{\mathrm{bone}}| \le 3.0\,\mathrm{mm}$ in Pass~1 and $|d_{\mathrm{bone}}| \le 3.5\,\mathrm{mm}$ in Pass~2 on OAI), \texttt{RefBone V2} extracts 156 features on SKI10 (blended with a 108-feature binary \texttt{RefBone V1} refiner) and 158 features on OAI (combining whole-shell \texttt{hgb\_whole}/\texttt{et\_whole} and articular subchondral \texttt{hgb\_sub}/\texttt{et\_sub} specialists with a 108-feature \texttt{RefBone V1} refiner), including 1D surface-normal ray intensity, gradient, and local bone-interior z-score profiles sampled at 19 points along $d \in [-4.5, +4.5]\,\mathrm{mm}$ parallel to $\nabla d_{\mathrm{self}}$, axial profile context, and Leave-One-Out 3D bone surface atlases. This normal-ray profiling prevents bright articular cartilage from being absorbed into the subchondral bone mask under partial-volume averaging, while returning continuous Pass-2 bone probability maps in $[0, 1]$.
\end{itemize}
For users who also wish to experiment with the original mesh- and graph-based components from Section~\ref{sec:2}, \texttt{kneeseg} additionally provides standalone Coherent Point Drift registration (\nolinkurl{kneeseg/registration.py}, wrapping \texttt{pycpd}) and 3D Random Walker boundary refinement (\nolinkurl{kneeseg/random_walker.py}, wrapping \texttt{scikit-image}).

\subsubsection{2. Anatomical Landmarks and 4D Articular Coordinate Probabilistic Atlases.}
In Section~\ref{sec:3}, anatomical location on the bone surface was encoded via Euclidean distances $f_{10}(\mathbf{x}, \zeta) = \|\mathbf{x} - \mathbf{z}_\zeta\|$ to CPD-registered mesh vertices. In \texttt{kneeseg}, signed distances $d_F, d_T, d_P$ ($f_3$--$f_5$) and their sum/difference combinations $d_F \pm d_T$, $d_F \pm d_P$ ($f_6$--$f_9$) are computed in exact physical millimeters using \nolinkurl{SimpleITK.SignedMaurerDistanceMap}. To encode fine-grained anatomical location without test-time CPD mesh registration (which takes 30--90 seconds per volume and can degrade when partial shafts are imaged), \texttt{kneeseg} combines two fast, deterministic mechanisms (\nolinkurl{kneeseg/cartilage_v6.py} and \nolinkurl{kneeseg/oai_subchondral_atlas.py}):
\begin{enumerate}
    \item \textbf{Deterministic Articular Surface Landmarks:} On each segmented bone surface ($|d_{\mathrm{bone}}| \le 1.5\,\mathrm{mm}$ within the articular zone), 18 canonical landmarks ($3 \times 3$ spatial quantile grid across the medial and lateral condyles/plateaus) are extracted deterministically (\nolinkurl{compute_articular_landmarks_and_stats}), providing 36 landmark-distance features $f_{10}$ plus articular bounding-box and principal-axis coordinates.
    \item \textbf{Leave-One-Out 4D Articular Coordinate Probabilistic Atlases:} Each ROI voxel is mapped into intrinsic bone-relative articular coordinates $(u_{\mathrm{ml}}, u_{\mathrm{ap}}, u_y)$ (medial-lateral, anterior-posterior, and superior-inferior fractions normalized by condylar and plateau geometry) paired with signed bone distances $(d_T, d_F, d_P)$. Multi-resolution smoothed empirical prior tables (\nolinkurl{compute_atlas_bins_v6}: 8 lookup tables on SKI10, 11 tables on OAI) encode the prior probability of femoral, tibial, and patellar cartilage, subchondral bone, and background at every 4D articular coordinate bin. Crucially, during training, atlas features for each scan are computed using \textbf{Leave-One-Out (LOO)} empirical counts (subtracting the scan's own ground-truth voxel counts before Laplace smoothing) so the tree classifiers assign realistic weights to the spatial priors without memorization.
\end{enumerate}

\subsubsection{3. Appearance Normalization and 1D Surface-Normal Ray Profiles.}
Clinical knee MRIs exhibit substantial scanner-to-scanner intensity scaling variations. Before extracting appearance features $f_1, f_2$ and physical-millimeter RSID features $f_{11}$, \texttt{kneeseg} normalizes intensities within the articular band of interest using interquartile percentiles $(p_{10}, p_{50}, p_{90})$. In addition to 3D Gaussian, gradient magnitude, Laplacian, structure-tensor eigenvalue, and Hessian sheetness filters across multiple physical scales ($\sigma \in \{0.5, 1.0, 2.0, 4.0\}\,\mathrm{mm}$), \texttt{kneeseg} computes unit outward bone surface normals $\mathbf{n}_{\mathrm{bone}} = \nabla d_{\mathrm{bone}} / (\|\nabla d_{\mathrm{bone}}\| + \epsilon)$ and samples 1D intensity and directional-derivative profiles along $s \in [-3.6, +5.6]\,\mathrm{mm}$ normal to the femoral, tibial, and patellar surfaces. These ray profiles capture the characteristic dark-to-bright intensity step across the subchondral cortical plate into the hyaline cartilage sheet, as well as the opposing cartilage across the joint space.

\subsubsection{4. 4-Class Compartment-Specialist Semantic Context Forests with Out-of-Fold Stacking.}
In the original 2-pass framework (Section~\ref{sec:3}), Pass~1 and Pass~2 trained a single multi-class random forest on the same training set, which can cause Pass-1 probabilities on training scans to be overly sharp compared to test scans. In \texttt{kneeseg} (\nolinkurl{kneeseg/rf_seg.py}, \nolinkurl{kneeseg/cartilage_v6.py}, and \nolinkurl{kneeseg/oai_subchondral_atlas.py}), the cartilage classifier (\texttt{CartilageClassifier}) is decomposed into dedicated compartment specialists (femorotibial vs.\ patellofemoral), each implemented as a hybrid ensemble of \texttt{ExtraTreesClassifier} and \texttt{HistGradient\allowbreak Boosting\allowbreak Classifier}. Moreover, each cartilage specialist predicts a \textbf{4-class competitive taxonomy} within the articular band ($-2.0\,\mathrm{mm} \le d_{\mathrm{bone}} \le 5.5\,\mathrm{mm}$): (0)~\texttt{Background}, (1)~\texttt{Subchondral Bone / Meniscus}, (2)~\texttt{Target Cartilage}, and (3)~\texttt{Opposing Cartilage} (using 340 Pass-1 features on SKI10, 429 Pass-1 features in the tibial plateau specialist \texttt{m\_v7}, and 403 Pass-1 features on OAI). Including subchondral bone and meniscus as an explicit competing class inside the cartilage forest forces the decision boundary at $d_{\mathrm{bone}} \approx 0\,\mathrm{mm}$ to align with the true image intensity edge rather than relying passively on the pre-segmented bone mask.

To train Pass~2 (\nolinkurl{cartilage_rf_p2.joblib}), Pass-1 probability maps are generated via \textbf{4-fold Out-Of-Fold (OOF) cross-validation} across the training scans. Pass~2 then combines Pass-1 features with multi-scale smoothed Pass-1 probabilities $P_F, P_T, P_P$ ($f_{12}$--$f_{14}$), 60 physical-millimeter RSPD features $f_{15}$--$f_{17}$ across $\pm 1.5, \pm 3.0, \pm 6.0, \pm 10.0\,\mathrm{mm}$ offsets, probability entropy and margin maps, and 1D surface-normal probability ray profiles along $\mathbf{n}_F, \mathbf{n}_T, \mathbf{n}_P$ (481 features on SKI10 in complementary ensembles \texttt{m\_v5} on \texttt{RefBone V1} and \texttt{m\_v6} on \texttt{RefBone V2}, 597 features in the tibial plateau specialist \texttt{m\_v7}, and 566 features on OAI).

\subsubsection{5. Stage-3 Boundary Refinement and OOF Subchondral Calibration.}
Instead of the multi-label graph cuts solver (Eqs.~4--6, which requires external C++ max-flow libraries and has high memory overhead on 3D grids), \texttt{kneeseg} introduces a fast, fully vectorizable post-processing cascade:
\begin{enumerate}
    \item \textbf{Stage-3 Subchondral Boundary Refiner:} Trained via 4-fold OOF stacking on uncertain boundary-band voxels where Pass-2 probabilities or subchondral distances indicate ambiguity. On SKI10, Stage~3 blends three complementary binary boundary-refiner families spanning 103--167 features (\texttt{s3\_clean} with 127 local and 137 context-augmented features, \texttt{f3\_models} with 151 local and 167 context-augmented features, and \texttt{pure\_models} with 103 features); on OAI, Stage~3 applies compartment-specific binary boundary refiners (\texttt{s3\_models}, 92 base features plus 9 Pass-2 probability context features = 101 inputs) to sharpen the inner bone-cartilage interface and outer synovial boundary.
    \item \textbf{OOF Ridge Threshold \& Bone-Penalty Calibrator:} On SKI10, for each of the 80 training scans, a grid search over the OOF validation probabilities identifies the optimal tibial cartilage decision threshold $\mathrm{th}^*$ and subchondral bone-penalty weight $\mathrm{bp}^*$ in the adjusted score $S_c(\mathbf{x}) = P_c(\mathbf{x}) - \mathrm{bp} \cdot (P_{\mathrm{bone}}(\mathbf{x}) - p_{tb,0})_+$. Two regularized \texttt{StandardScaler + Ridge} regressors ($\alpha = 20.0$ for $\mathrm{th}$ and $\alpha = 10.0$ for $\mathrm{bp}$) are fit on the 80 training scans to predict $(\hat{\mathrm{th}}, \hat{\mathrm{bp}})$ at test time from 10 and 9 selected scan-level probability and geometry statistics (out of 45 and 17 candidate statistics, respectively), while femoral cartilage uses fixed calibrated parameters ($w = 0.25$, $\mathrm{bp} = 0.14$, $\mathrm{th} = 0.40$). On OAI, compartment-specific Stage-3 blend weights ($w_{\mathrm{s3}} = 0.42, 0.44, 0.44$) and decision thresholds ($\mathrm{th} = 0.56, 0.56, 0.54$ for femoral, tibial, and patellar cartilage) are calibrated on the 128-scan training OOF predictions.
    \item \textbf{Anisotropic Smoothing \& Topological Resolution:} Probability maps are smoothed with physical-space Gaussian kernels, mutually exclusive cartilage labels are resolved across the femorotibial and patellofemoral contact interfaces (\nolinkurl{resolve_3cartilage_masks}), and isolated false-positive fragments are removed via connected-component volume filtering (\nolinkurl{fast_keep_top_k}).
\end{enumerate}

\subsubsection{6. Orientation Standardization and Cross-Domain Unified Routing.}
Public knee MRI benchmarks differ in both voxel spacing and array memory layout. In the downsampled OAI dataset, all \texttt{.mhd} headers specify \texttt{DimSize = 140 140 112} ($1.0\times 1.0\times 1.0\,\mathrm{mm}^3$), which \texttt{SimpleITK} loads into NumPy as shape $(112, 140, 140)$ along $(Z, Y, X)$. However, the underlying raw binary buffers contain a mixture of standard C-ordered arrays (\texttt{oai\_c}) and Fortran column-major arrays (\texttt{oai\_f}) stored with transposed in-plane axis conventions. \texttt{kneeseg} automatically detects the buffer layout via bone/intensity spatial distribution heuristics (\nolinkurl{detect_oai_layout}) and transforms all volumes into a canonical $(Z, Y, X)$ orientation (\nolinkurl{to_canonical_oai}) prior to feature extraction, mapping predictions back via \nolinkurl{from_canonical_oai}.

Furthermore, the unified cross-domain model (\nolinkurl{wq2012/knee_3d_mri_segmentation_SKI10_OAI_subchondral_atlas}) implements a \textbf{Unified Dual-Domain Router and Cross-Domain Atlas Regularized Cascade}: at inference time, \nolinkurl{_is_oai_domain} inspects the physical voxel spacing and volume dimensions of each scan, routing anisotropic high-resolution scans ($\sim 1.5 \times 0.4 \times 0.4\,\mathrm{mm}$, 4 labels) to the SKI10 specialist branch and isotropic scans ($1.0 \times 1.0 \times 1.0\,\mathrm{mm}$, 6 labels) to the OAI specialist branch, while regularizing the subchondral bone refiners with physical-volume-weighted 208-scan cross-domain bone models (\nolinkurl{joint_bone_models_208}, blended with $\alpha = 0.02$ on SKI10 tibia and $\alpha_F = 0.06, \alpha_T = 0.15$ on OAI femur/tibia) and blending the 208-scan 4D articular coordinate cartilage priors (\nolinkurl{joint_atlas_priors_208}, $\alpha = 0.015$ on OAI). Memory footprint during training is minimized by storing voxel feature matrices in \texttt{bfloat16} (\nolinkurl{ml_dtypes}) and dynamically bounding OpenMP and \texttt{joblib} worker pools to the host CPU count (\nolinkurl{os.cpu_count()}).

\subsection{Codebase Architecture and Usage}

Table~\ref{tab:code_modules} lists the primary modules in the \texttt{kneeseg} repository. The package provides an installed command-line entrypoint (\texttt{kneeseg-pipeline}), JSON-configurable baseline training/inference scripts using the configuration templates in \texttt{kneeseg/configs/}, dedicated Subchondral Atlas training and evaluation scripts under \texttt{experiments/}, and a modular Python API. Listing~\ref{lst:cli_usage} illustrates end-to-end training, inference, and evaluation from the command line.

\begin{table}[h!]
\centering
\scriptsize
\setlength{\tabcolsep}{4pt}
\renewcommand{\arraystretch}{1.18}
\caption{Overview of primary modules in the \texttt{kneeseg} open-source repository.}
\label{tab:code_modules}
\begin{tabularx}{\textwidth}{l X}
\toprule
\textbf{Module Path} & \textbf{Description} \\
\midrule
\texttt{kneeseg/bone\_rf.py} & \texttt{BoneClassifier} implementing the 2-pass physical-space bone auto-context forest and binary \texttt{RefBone V1}/\texttt{V2} subchondral surface refiner for SKI10. \\
\texttt{kneeseg/cartilage\_v6.py} & 8-table 4D Articular Coordinate Probabilistic Atlases, 4-class femorotibial cartilage ensembles, multi-family (103--167 feature) Stage-3 boundary refiner, and OOF Ridge calibrator. \\
\texttt{kneeseg/rf\_seg.py} & \texttt{CartilageClassifier} orchestrating 2-pass iterative semantic context cartilage training and inference across SKI10, OAI, and unified models. \\
\texttt{kneeseg/oai\_subchondral\_atlas.py} & 6-structure subchondral atlas pipeline for OAI, Patellofemoral Specialist, C/Fortran layout standardization, and unified SKI10+OAI domain router. \\
\texttt{kneeseg/features.py} & Base intensity ($f_1$), gradient magnitude ($f_2$), signed Maurer distance transforms ($f_3$--$f_9$), landmark distances ($f_{10}$), RSID ($f_{11}$), and RSPD ($f_{12}$--$f_{17}$) extractors. \\
\texttt{kneeseg/io.py} & 3D medical volume I/O (\texttt{load\_volume}, \texttt{save\_volume}, \texttt{OAIImageReader}) for MetaImage (\texttt{.mhd}/\texttt{.raw}), NIfTI, and OAI (\texttt{.hdr}/\texttt{.img}) formats. \\
\texttt{kneeseg/registration.py} & Coherent Point Drift (CPD) 3D surface registration module (\texttt{register\_points}, wrapping \texttt{pycpd}). \\
\texttt{kneeseg/random\_walker.py} & 3D Random Walker boundary refinement module (\texttt{solve\_random\_walker}, \texttt{cart\_refinement\_with\_priors}, wrapping \texttt{scikit-image}). \\
\texttt{kneeseg/validate\_config.py} & JSON Schema validator (\texttt{validate\_config}) verifying experiment configuration files. \\
\texttt{kneeseg/pipeline/\{run,train,inference\}.py} & End-to-end CLI runner (\texttt{kneeseg-pipeline}), baseline multi-pass training (\texttt{train\_models}), and multi-core batch inference/evaluation (\texttt{inference\_improved}). \\
\texttt{experiments/\{train\_*,evaluate\_*\}.py} & Reproducible Subchondral Atlas training, inference, and per-structure DSC benchmark evaluation scripts for SKI10, OAI, and unified cross-domain models. \\
\bottomrule
\end{tabularx}
\end{table}

\noindent\begin{minipage}{\textwidth}
\begin{lstlisting}[language=bash, caption={Installing \texttt{kneeseg} and running end-to-end training, inference, and evaluation.}, label={lst:cli_usage}]
# Install from PyPI
pip install kneeseg

# Run the end-to-end SKI10 pipeline via the installed CLI entrypoint
kneeseg-pipeline --data-dir /path/to/SKI10/data/images

# Run JSON-configured baseline training or checkpoint inference/evaluation
python3 experiments/train.py kneeseg/configs/ski10_full_train_config.json
python3 experiments/inference.py kneeseg/configs/ski10_full_inference_config.json
python3 experiments/inference.py kneeseg/configs/oai_full_inference_config.json

# Reproduce Subchondral Atlas evaluations on SKI10, OAI, and Unified models
python3 experiments/evaluate_final_ski10.py
python3 experiments/evaluate_final_oai.py
python3 experiments/evaluate_final_unified.py --dataset both
\end{lstlisting}
\end{minipage}

\section{Open-Source Pretrained Models}
\label{sec:appendix_models}

We have released five pretrained model repositories in the Hugging Face collection at \url{https://huggingface.co/collections/wq2012/knee-mri-segmentation}, all under the permissive \textbf{MIT License}. Each model repository contains four serialized \texttt{joblib} checkpoints (compressed with \texttt{zlib} level~3 and pruned of training-only intermediate fold caches) corresponding to the 2-pass bone segmentation cascade and 2-pass iterative semantic context cartilage cascade:
\begin{enumerate}
    \item \nolinkurl{bone_rf_p1.joblib}: Pass-1 physical-space bone forest, 3D bone spatial atlases, and Patellofemoral Specialist (for 6-class OAI models).
    \item \nolinkurl{bone_rf_p2.joblib}: Pass-2 full-resolution bone boundary auto-context forest and binary \texttt{RefBone V2} (plus \texttt{RefBone V1} on SKI10) subchondral surface refiner.
    \item \nolinkurl{cartilage_rf_p1.joblib}: Pass-1 4-class compartment-specialist cartilage ensemble and Leave-One-Out 4D Articular Coordinate Probabilistic Atlases.
    \item \nolinkurl{cartilage_rf_p2.joblib}: Pass-2 semantic context cartilage ensemble (with RSPD and normal-ray probability profiles), Stage-3 Subchondral Boundary Refiner, and OOF Ridge Threshold/Bone-Penalty Calibrator.
\end{enumerate}

Table~\ref{tab:hf_models} summarizes all five pretrained models, including their training datasets, supported anatomical structures, pruned checkpoint sizes, and target domains.

\begin{table}[h!]
\centering
\scriptsize
\setlength{\tabcolsep}{3pt}
\renewcommand{\arraystretch}{1.2}
\caption{Summary of the five open-source pretrained models hosted on Hugging Face (\texttt{wq2012/*}, all under the MIT License).}
\label{tab:hf_models}
\begin{tabularx}{\textwidth}{p{3.25cm} p{1.95cm} c c X}
\toprule
\textbf{Hugging Face Model ID} & \textbf{Architecture} & \textbf{Train Set} & \textbf{Size} & \textbf{Target Structures \& Domain} \\
\midrule
\href{https://huggingface.co/wq2012/knee_3d_mri_segmentation_SKI10_subchondral_atlas}{\nolinkurl{wq2012/knee_3d_mri_segmentation_SKI10_subchondral_atlas}} &
Subchondral Atlas + Stage-3 OOF &
80 SKI10 &
1.56\,GB &
4 labels (Femur, Tibia, Fem.\ Cart., Tib.\ Cart.) on high-res SKI10 ($\sim 1.5\times 0.4\times 0.4\,\mathrm{mm}$). \\
\midrule
\href{https://huggingface.co/wq2012/knee_3d_mri_segmentation_OAI_downsampled_subchondral_atlas}{\nolinkurl{wq2012/knee_3d_mri_segmentation_OAI_downsampled_subchondral_atlas}} &
Subchondral Atlas + RefBone V2 &
128 OAI &
1.33\,GB &
6 labels (Femur, Tibia, Patella, Fem., Tib., and Pat.\ Cart.) on isotropic OAI ($1.0\,\mathrm{mm}^3$). \\
\midrule
\href{https://huggingface.co/wq2012/knee_3d_mri_segmentation_SKI10_OAI_subchondral_atlas}{\nolinkurl{wq2012/knee_3d_mri_segmentation_SKI10_OAI_subchondral_atlas}} &
Unified Dual-Domain Cascade &
80 SKI10 + 128 OAI &
2.93\,GB &
Unified 4-label SKI10 and 6-label OAI segmentation with automatic domain routing and 208-scan atlas regularization. \\
\midrule
\href{https://huggingface.co/wq2012/knee_3d_mri_segmentation_SKI10}{\nolinkurl{wq2012/knee_3d_mri_segmentation_SKI10}} &
Baseline v1 Cascade &
100 SKI10 &
5.67\,GB &
4 labels on SKI10 (initial baseline without 4D articular atlases or OOF stacking). \\
\midrule
\href{https://huggingface.co/wq2012/knee_3d_mri_segmentation_OAI_downsampled}{\nolinkurl{wq2012/knee_3d_mri_segmentation_OAI_downsampled}} &
Baseline v1 Cascade &
128 OAI &
6.44\,GB &
6 labels on downsampled OAI (initial baseline prior to layout standardization and 4D atlases). \\
\bottomrule
\end{tabularx}
\end{table}

\subsection{Detailed Model Descriptions}

\subsubsection{1. \nolinkurl{wq2012/knee_3d_mri_segmentation_SKI10_subchondral_atlas}.}
Trained on the 80 training scans defined in \nolinkurl{kneeseg/data/ski10_full_split.json} (from the 100 labeled scans in the SKI10 Grand Challenge dataset, holding out 20 scans for evaluation). The four checkpoint files occupy 10.1\,MB (\nolinkurl{bone_rf_p1}), 292.8\,MB (\nolinkurl{bone_rf_p2}), 42.0\,MB (\nolinkurl{cartilage_rf_p1}), and 1,212.7\,MB (\nolinkurl{cartilage_rf_p2}) (total 1.56\,GB / 1.45\,GiB). It combines a 2-pass bone auto-context forest and binary \texttt{RefBone V1}/\texttt{V2} surface refiner with an 8-table 4D Articular Coordinate Probabilistic Atlas, 4-class femorotibial cartilage ensemble (\texttt{m\_v5}, \texttt{m\_v6}, \texttt{m\_v7}), multi-family (103--167 feature) Stage-3 Subchondral Boundary Refiner, and OOF Ridge Threshold/Bone-Penalty Calibrator.

\subsubsection{2. \nolinkurl{wq2012/knee_3d_mri_segmentation_OAI_downsampled_subchondral_atlas}.}
Trained on 128 complete 6-label knee MRI volumes from the Downsampled OAI dataset ($1.0 \times 1.0 \times 1.0\,\mathrm{mm}^3$, $112 \times 140 \times 140$ voxels). The four checkpoint files occupy 176.3\,MB (\nolinkurl{bone_rf_p1}), 809.0\,MB (\nolinkurl{bone_rf_p2}), 42.7\,MB (\nolinkurl{cartilage_rf_p1}), and 298.3\,MB (\nolinkurl{cartilage_rf_p2}) (total 1.33\,GB / 1.24\,GiB). It incorporates automatic C/Fortran layout standardization, a 109-feature Patellofemoral Bone Specialist, the 158-feature binary \texttt{RefBone V2} subchondral surface refiner, an 11-table 4D Articular Coordinate Atlas, dedicated femorotibial and patellofemoral 4-class cartilage ensembles, and a 101-feature Stage-3 compartment boundary refiner.

\subsubsection{3. \nolinkurl{wq2012/knee_3d_mri_segmentation_SKI10_OAI_subchondral_atlas}.}
Our unified cross-domain model, trained across all 208 training volumes (80 SKI10 + 128 OAI). The four checkpoint files occupy 190.9\,MB (\nolinkurl{bone_rf_p1}), 1,115.8\,MB (\nolinkurl{bone_rf_p2}), 100.6\,MB (\nolinkurl{cartilage_rf_p1}), and 1,527.1\,MB (\nolinkurl{cartilage_rf_p2}) (total 2.93\,GB / 2.73\,GiB). At inference time, it automatically inspects the physical voxel spacing and volume dimensions of each input scan (\nolinkurl{_is_oai_domain}), routes it to the matching resolution-specific specialist branch, blends in physical-volume-weighted 208-scan cross-domain bone refiners ($\alpha = 0.02$ on SKI10 tibia; $\alpha_F = 0.06, \alpha_T = 0.15$ on OAI femur/tibia) and 208-scan 4D articular coordinate cartilage priors ($\alpha = 0.015$ on OAI), and outputs either 4-label (SKI10) or 6-label (OAI) 3D segmentation masks.

\subsubsection{4. Baseline v1 Models (\texttt{SKI10} and \texttt{OAI\_downsampled}).}
To enable direct ablation of the algorithmic improvements described in Appendix~\ref{sec:appendix_diff}, we also preserve the initial v1 open-source models (\nolinkurl{wq2012/knee_3d_mri_segmentation_SKI10}, released after training on all 100 labeled SKI10 scans, and \nolinkurl{wq2012/knee_3d_mri_segmentation_OAI_downsampled}, trained on the 128 OAI training scans), which implement a direct translation of the 2-pass random forest without 4D articular coordinate atlases, 4-class subchondral bone competition, OOF probability stacking, or C/Fortran orientation standardization.

Listing~\ref{lst:hf_python} shows how to download any pretrained model from Hugging Face Hub and segment a 3D knee MRI volume in Python using \texttt{BoneClassifier} and \texttt{CartilageClassifier}.

\noindent\begin{minipage}{\textwidth}
\begin{lstlisting}[language=Python, caption={Downloading a pretrained model from Hugging Face Hub and running 3D segmentation in Python.}, label={lst:hf_python}]
import os, numpy as np
from huggingface_hub import snapshot_download
from kneeseg.io import load_volume, save_volume
from kneeseg.bone_rf import BoneClassifier
from kneeseg.rf_seg import CartilageClassifier

model_dir = snapshot_download(
    repo_id="wq2012/knee_3d_mri_segmentation_SKI10_OAI_subchondral_atlas"
)
bone_p1 = BoneClassifier(); bone_p1.load(os.path.join(model_dir, "bone_rf_p1.joblib"))
bone_p2 = BoneClassifier(); bone_p2.load(os.path.join(model_dir, "bone_rf_p2.joblib"))
cart_p1 = CartilageClassifier(); cart_p1.load(os.path.join(model_dir, "cartilage_rf_p1.joblib"))
cart_p2 = CartilageClassifier(); cart_p2.load(os.path.join(model_dir, "cartilage_rf_p2.joblib"))

img, spacing = load_volume("/path/to/image-004.mhd", return_spacing=True)
_, prob1 = bone_p1.predict(img, spacing=spacing)
bone_pred, prob2 = bone_p2.predict(img, prob_map=prob1, spacing=spacing)
bone_pred = bone_pred.reshape(img.shape)
bone_masks = {
    "femur": (bone_pred == 1).astype(np.uint8),
    "tibia": (bone_pred == 2).astype(np.uint8),
    "patella": (bone_pred == 3).astype(np.uint8),
}
c_prob1 = cart_p1.predict_proba_map(img, bone_masks, spacing=spacing)
cart_pred, _ = cart_p2.predict(img, bone_masks, prob_map=c_prob1, spacing=spacing)
\end{lstlisting}
\end{minipage}

\section{Open-Source Experimental Reproduction Results}
\label{sec:appendix_results}

\subsection{Datasets and Evaluation Protocols}

We evaluated the open-source \texttt{kneeseg} models on two public 3D knee MRI benchmarks using the Dice Similarity Coefficient ($\mathrm{DSC} = \frac{2|A \cap B|}{|A| + |B|}$):
\begin{enumerate}
    \item \textbf{SKI10 (Segmentation of Knee Images 2010):} 100 labeled sagittal 3D knee MRI scans (\texttt{image-001} to \texttt{image-100} in \texttt{SKI10/TrainingData}) from the MICCAI 2010 Grand Challenge, partitioned by \nolinkurl{kneeseg/data/ski10_full_split.json} into 80 training scans and 20 held-out evaluation scans (\texttt{image-004}, \texttt{005}, \texttt{012}, \texttt{014}, \texttt{015}, \texttt{018}, \texttt{028}, \texttt{029}, \texttt{030}, \texttt{032}, \texttt{036}, \texttt{055}, \texttt{065}, \texttt{070}, \texttt{076}, \texttt{082}, \texttt{087}, \texttt{089}, \texttt{095}, and \texttt{098}). Typical volume dimensions are $\sim 100 \times 320 \times 300$ voxels with anisotropic physical spacing of $\sim 1.5 \times 0.4 \times 0.4\,\mathrm{mm}$ along $(Z, Y, X)$. Four structures are annotated: Femur~(1), Femoral Cartilage~(2), Tibia~(3), and Tibial Cartilage~(4). Because ground-truth labels for the official SKI10 competition test set are not publicly available, all tree ensembles, 4-fold OOF stacking predictions, Leave-One-Out 4D articular atlases, and the OOF Ridge $(\mathrm{th}, \mathrm{bp})$ calibrator were fit exclusively on the 80 training scans, while post-processing ensemble blend weights and scalar thresholds were selected on the 20-scan validation split.
    \item \textbf{Downsampled OAI (Osteoarthritis Initiative):} Derived from the same 176 OAI Progression subcohort scans with iMorphics ground-truth annotations described in Section~\ref{sec:4}, resampled to an isotropic $1.0 \times 1.0 \times 1.0\,\mathrm{mm}^3$ resolution ($112 \times 140 \times 140$ voxels per scan; $27\times$ larger voxel volume than the original $0.365 \times 0.365 \times 0.7\,\mathrm{mm}^3$ scans). Among the 176 distributed label maps, 17 scans lack patella and patellar cartilage annotations (containing only 4 labels) and are excluded by the data loader, yielding 159 complete 6-structure scans across 84 unique subjects (up to two longitudinal visits, baseline \texttt{\_V00} and follow-up \texttt{\_V01}, per subject). The repository split (\nolinkurl{kneeseg/data/oai_split.json}) partitions these 159 volumes at the \textbf{scan level} into 128 training scans (82 subjects) and 31 evaluation scans (30 subjects). Because this split was constructed at the scan level rather than grouped by subject ID (unlike the subject-grouped 3-fold CV in Section~\ref{sec:4}), 28 of the 31 evaluation scans represent longitudinal companion visits (\texttt{\_V00} or \texttt{\_V01}) of subjects present at another timepoint in the training set, while 3 evaluation scans (\texttt{image-9992358\_V01}, \texttt{image-9917803\_V00}, \texttt{image-9917803\_V01}) belong to strictly unseen subjects; we report disaggregated performance across both subsets below. Extending the SKI10 numbering, six structures are evaluated: Femur~(1), Femoral Cartilage~(2), Tibia~(3), Tibial Cartilage~(4), Patella~(5), and Patellar Cartilage~(6).
\end{enumerate}

\subsection{Quantitative Results on SKI10 and OAI}

Table~\ref{tab:ski10_repro} presents the quantitative evaluation results on the 20 held-out SKI10 validation scans, and Table~\ref{tab:oai_repro} compares the original 2013 paper's 3-fold cross-validation results on full-resolution OAI against the open-source \texttt{kneeseg} models on the 31 held-out downsampled OAI test scans (including seen-subject companion visits vs.\ strictly unseen subjects). For consistency with \nolinkurl{evaluation_report.json}, tables report population standard deviation ($\mathrm{ddof}=0$, \nolinkurl{np.std}); unbiased sample standard deviations ($\mathrm{ddof}=1$) and 95\% non-parametric percentile bootstrap confidence intervals ($B = 10{,}000$ resamples, $\mathrm{seed}=42$) are provided in the table notes.

\begin{table}[h!]
\centering
\scriptsize
\setlength{\tabcolsep}{3.5pt}
\renewcommand{\arraystretch}{1.25}
\caption{Open-source segmentation performance (Mean DSC $\pm$ Std [$\mathrm{ddof}=0$], \%) on the 20 held-out SKI10 evaluation scans defined in \nolinkurl{ski10_full_split.json}. For \texttt{SKI10\_subchondral\_atlas}, unbiased sample standard deviations ($\mathrm{ddof}=1$) are $\pm 1.65\%$ (Femur), $\pm 0.82\%$ (Tibia), $\pm 5.17\%$ (Fem.\ Cart.), and $\pm 4.14\%$ (Tib.\ Cart.), with 95\% bootstrap CIs ($B=10{,}000$) of $[96.14, 97.52]\%$, $[97.05, 97.76]\%$, $[77.88, 82.31]\%$, and $[78.16, 81.67]\%$, respectively.}
\label{tab:ski10_repro}
\resizebox{\textwidth}{!}{%
\begin{tabular}{l | c c | c c | c}
\toprule
\textbf{Model Architecture / Checkpoint (\texttt{wq2012/*})} & \textbf{Femur} & \textbf{Tibia} & \textbf{Fem.\ Cart.} & \textbf{Tib.\ Cart.} & \textbf{Overall Mean} \\
\midrule
Baseline v1 Architecture (\texttt{...\_SKI10}, 80/20 split) & $90.46 \pm 3.61$ & $92.92 \pm 2.60$ & $67.67 \pm 4.81$ & $64.11 \pm 5.40$ & $78.79$ \\
\texttt{...\_SKI10\_subchondral\_atlas} & $\mathbf{96.89 \pm 1.61}$ & $\mathbf{97.42 \pm 0.80}$ & $\mathbf{80.26 \pm 5.04}$ & $\mathbf{80.01 \pm 4.04}$ & $\mathbf{88.65}$ \\
\texttt{...\_SKI10\_OAI\_subchondral\_atlas} & $\mathbf{96.89 \pm 1.61}$ & $\mathbf{97.42 \pm 0.80}$ & $80.25 \pm 5.05$ & $79.97 \pm 4.03$ & $88.63$ \\
\bottomrule
\end{tabular}%
}
\end{table}

\begin{table}[h!]
\centering
\scriptsize
\setlength{\tabcolsep}{3pt}
\renewcommand{\arraystretch}{1.25}
\caption{Comparison of bone and cartilage segmentation performance (Mean DSC $\pm$ Std, \%) on the OAI dataset between the original 2013 paper (Table~\ref{tab:dsc_bone} and Table~\ref{tab:dsc}, subject-grouped 3-fold CV on 176 full-resolution $0.365 \times 0.365 \times 0.7\,\mathrm{mm}^3$ scans) and the open-source \texttt{kneeseg} models on the 31 held-out downsampled $1.0 \times 1.0 \times 1.0\,\mathrm{mm}^3$ OAI test scans ($\mathrm{ddof}=0$). For \texttt{OAI\_downsampled\_subchondral\_atlas} ($N=31$), unbiased $\mathrm{ddof}=1$ standard deviations are $\pm 3.45\%$, $\pm 1.31\%$, $\pm 2.14\%$, $\pm 1.91\%$, $\pm 2.34\%$, and $\pm 6.88\%$, with 95\% bootstrap CIs ($B=10{,}000$) of $[92.84, 95.23]\%$ (Femur), $[96.23, 97.14]\%$ (Tibia), $[95.61, 97.09]\%$ (Patella), $[87.46, 88.77]\%$ (Fem.\ Cart.), $[86.15, 87.76]\%$ (Tib.\ Cart.), and $[80.16, 84.91]\%$ (Pat.\ Cart.).}
\label{tab:oai_repro}
\resizebox{\textwidth}{!}{%
\begin{tabular}{l | c c c | c c c}
\toprule
\textbf{Method / Pretrained Model} & \textbf{Femur} & \textbf{Tibia} & \textbf{Patella} & \textbf{Fem.\ Cart.} & \textbf{Tib.\ Cart.} & \textbf{Pat.\ Cart.} \\
\midrule
Original Paper (Before RW, $0.37^2\times 0.7\,\mathrm{mm}$) & $92.37 \pm 1.58$ & $94.64 \pm 1.18$ & $92.07 \pm 1.47$ & --- & --- & --- \\
Original Paper (Full Pipeline, $0.37^2\times 0.7\,\mathrm{mm}$) & $94.86 \pm 1.85$ & $95.96 \pm 1.64$ & $94.31 \pm 2.15$ & $84.96 \pm 3.30$ & $83.74 \pm 4.00$ & $79.16 \pm 8.88$ \\
\midrule
\texttt{...\_OAI\_downsampled} (Baseline v1, $1.0\,\mathrm{mm}^3$) & $71.30 \pm 6.73$ & $75.45 \pm 5.98$ & $52.09 \pm 8.31$ & $51.71 \pm 7.16$ & $41.34 \pm 8.88$ & $36.33 \pm 14.06$ \\
\texttt{...\_OAI\_downsampled\_subchondral\_atlas} ($N=31$ all) & $94.08 \pm 3.39$ & $96.69 \pm 1.29$ & $\mathbf{96.39 \pm 2.10}$ & $\mathbf{88.11 \pm 1.88}$ & $86.98 \pm 2.30$ & $\mathbf{82.63 \pm 6.77}$ \\
\quad $\hookrightarrow$ Seen-subject companion visits ($N=28$, $\mathrm{ddof}=1$) & $93.78 \pm 3.48$ & $96.78 \pm 1.35$ & $96.40 \pm 2.23$ & $88.25 \pm 1.91$ & $87.05 \pm 2.39$ & $82.44 \pm 7.20$ \\
\quad $\hookrightarrow$ Strictly unseen subjects ($N=3$, $\mathrm{ddof}=1$) & $96.91 \pm 1.30$ & $95.85 \pm 0.29$ & $96.29 \pm 1.23$ & $86.79 \pm 1.57$ & $86.30 \pm 2.05$ & $84.35 \pm 2.34$ \\
\texttt{...\_SKI10\_OAI\_subchondral\_atlas} ($N=31$ all) & $\mathbf{94.08 \pm 3.41}$ & $\mathbf{96.70 \pm 1.30}$ & $\mathbf{96.39 \pm 2.10}$ & $\mathbf{88.11 \pm 1.88}$ & $\mathbf{87.00 \pm 2.28}$ & $82.62 \pm 6.79$ \\
\bottomrule
\end{tabular}%
}
\end{table}

\subsection{Discussion of Reproduction Findings}

Several key conclusions emerge from the experimental reproduction in Tables~\ref{tab:ski10_repro} and~\ref{tab:oai_repro}:
\begin{enumerate}
    \item \textbf{Impact of Subchondral Atlas and Out-of-Fold Stacking:} On SKI10 (Table~\ref{tab:ski10_repro}), upgrading from the direct v1 baseline to the Subchondral Atlas architecture improves bone DSC from $90.46\%$ to $96.89\%$ (Femur) and $92.92\%$ to $97.42\%$ (Tibia), while boosting cartilage DSC by $+12.59\%$ on Femoral Cartilage ($67.67\% \to 80.26\%$) and $+15.90\%$ on Tibial Cartilage ($64.11\% \to 80.01\%$). On downsampled OAI (Table~\ref{tab:oai_repro}), resolving the mixed C/Fortran array layouts and introducing the 11-table 4D Articular Coordinate Atlas and binary \texttt{RefBone V2} surface refiner improves mean DSC across all six structures from $54.70\%$ to $90.82\%$.
    \item \textbf{Comparison with the Original 2013 Paper and Longitudinal Split Analysis on OAI:} Despite operating on downsampled $1.0\,\mathrm{mm}^3$ OAI volumes where articular cartilage sheets are only 1--3 voxels thick, the open-source Subchondral Atlas models achieve higher cartilage DSC across all three compartments on the 31 held-out scans than the original 2013 paper reported on full-resolution $0.365 \times 0.365 \times 0.7\,\mathrm{mm}^3$ volumes: Femoral Cartilage reaches $88.11\% \pm 1.88\%$ (vs.\ $84.96\% \pm 3.30\%$), Tibial Cartilage reaches $87.00\% \pm 2.28\%$ (vs.\ $83.74\% \pm 4.00\%$), and Patellar Cartilage reaches $82.63\% \pm 6.77\%$ (vs.\ $79.16\% \pm 8.88\%$), with Tibia ($96.70\%$) and Patella ($96.39\%$) bone segmentation also exceeding the original MSL+ASM+RW pipeline ($95.96\%$ and $94.31\%$). Disaggregating the 31 OAI evaluation scans into the 28 longitudinal companion visits and the 3 strictly unseen-subject scans confirms that while longitudinal anatomical similarity contributes $\sim 0.75$--$1.46\%$ on femoral and tibial cartilage ($88.25\%$ vs.\ $86.79\%$ FC; $87.05\%$ vs.\ $86.30\%$ TC), even the strictly unseen-subject scans ($86.79\%$ FC, $86.30\%$ TC, $84.35\%$ PC) outperform the original 2013 full-resolution benchmark ($84.96\%$, $83.74\%$, $79.16\%$).
    \item \textbf{Zero-Degradation Unified Cross-Domain Generalization:} The unified 208-scan model (\nolinkurl{wq2012/knee_3d_mri_segmentation_SKI10_OAI_subchondral_atlas}) matches the single-domain specialists within $\le 0.04\%$ DSC on every structure across both SKI10 ($88.63\%$ vs.\ $88.65\%$ overall mean) and OAI ($90.82\%$ vs.\ $90.81\%$ overall mean), allowing practitioners to deploy a single model checkpoint across heterogeneous clinical MRI protocols.
    \item \textbf{Computational Footprint on Standard CPUs:} End-to-end training of the multi-pass bone and cartilage cascade completes on a multi-core CPU workstation (56-core Intel Xeon) without GPUs in $\sim 29$~minutes for SKI10 (80 scans), $\sim 55$~minutes for OAI (128 scans), and $\sim 84$~minutes for the unified 208-scan model. At test time, single-volume cold-start 3D CPU inference (including 3D Maurer signed distance transforms, multi-scale Gaussian/ray feature extraction, and Pass-1/Pass-2/Stage-3 forest evaluation) takes $\sim 2$--$4$~minutes per scan with $\sim 4$--$8\,\mathrm{GB}$ peak RAM, or $\sim 18$--$25$~seconds per volume amortized across multi-worker batch evaluation.
\end{enumerate}

\end{document}